%% file: acl_latex.tex
\pdfoutput=1

\documentclass[11pt]{article}

\usepackage{authblk}
\usepackage[final]{acl}

\usepackage{times}
\usepackage{latexsym}
\usepackage{graphicx}
\usepackage{amssymb}

\usepackage{physics,amsmath}
\usepackage{cleveref}
\usepackage{enumitem}
\usepackage{booktabs}

\usepackage{textcomp, xspace}
\usepackage{pbox}
\usepackage{array}
\usepackage{multirow}
\usepackage{xcolor}
\usepackage[normalem]{ulem}
\useunder{\uline}{\ul}{}
\usepackage{rotating}
\usepackage{array}
\usepackage{listings}
\usepackage{adjustbox}

\crefformat{section}{\S#2#1#3} 
\crefformat{subsection}{\S#2#1#3}
\crefformat{subsubsection}{\S#2#1#3}
\usepackage{arydshln}
\usepackage[most]{tcolorbox}

\usepackage[T1]{fontenc}

\usepackage[utf8]{inputenc}

\usepackage{microtype}

\usepackage{inconsolata}

\usepackage{graphicx}

\newcommand{\methodFull}{\textsc{Narrative-of-Thought}\xspace}
\newcommand{\method}{\textsc{NoT}\xspace}

\newcommand{\nop}[1]{}

\newcommand{\red}[1]{\colorbox{magenta!45}{#1}}
\newcommand{\blue}[1]{\colorbox{cyan!45}{#1}}
\newcommand{\purple}[1]{\colorbox{violet!45}{#1}}

\title{\methodFull: Improving Temporal Reasoning of\\ Large Language Models via Recounted Narratives}

\author[1]{\textbf{Xinliang Frederick Zhang}}
\author[2]{\textbf{Nick Beauchamp}}
\author[1]{\textbf{Lu Wang}}

\affil[1]{Computer Science and Engineering, University of Michigan, Ann Arbor, MI}
\affil[2]{Department of Political Science, Northeastern University, Boston, MA}

\affil[ ]{$^1$\{\texttt{xlfzhang,wangluxy\}@umich.edu}, $^2$\texttt{n.beauchamp@northeastern.edu}}

\begin{document}
\maketitle
\input{0_abstract}

\input{1_introduction}
\input{2_related_word}

\input{3_method}

\input{4_experiment}
\input{5_results_analysis}

\input{6_conclusion}

\section*{Acknowledgments}
This work is supported in part through Air Force Office of Scientific Research under grant FA9550-22-1-0099, and computational resources and services provided by Advanced Research Computing (ARC), a division of Information and Technology Services (ITS) at the University of Michigan, Ann Arbor. We thank ARR reviewers for their valuable feedback.

\input{Limitations}

\clearpage

\clearpage
\setcounter{table}{0}
\setcounter{figure}{0}
\renewcommand{\thefigure}{A\arabic{figure}}
\renewcommand{\thetable}{A\arabic{table}}

\appendix
\input{Appendix}

\end{document}

%% file: 0_abstract.tex
\begin{abstract}
Reasoning about time and temporal relations is an integral aspect of human cognition, essential for perceiving the world and navigating our experiences. 
Though large language models (LLMs) 
have demonstrated impressive performance in many reasoning tasks, temporal reasoning remains challenging due to its intrinsic complexity. 
In this work, we first study an essential task of temporal reasoning---temporal graph generation, to unveil LLMs' inherent, global reasoning capabilities. 
We show that this task presents great challenges even for the most powerful LLMs, such as GPT-3.5/4. We also notice a significant performance gap by small models ($<10B$) that lag behind LLMs by $50\%$. 
Next, we study how to close this gap with a budget constraint, e.g., not using model finetuning. 
We propose a new prompting technique tailored for temporal reasoning, \methodFull (\method), that first converts the events set to a Python class, then prompts a small model to generate a temporally grounded narrative, guiding the final generation of a temporal graph. Extensive experiments showcase the efficacy of \method in improving various metrics. 
Notably, \method attains the highest F1 on the Schema-11 evaluation set, while securing an overall F1 on par with GPT-3.5.
\method also achieves the best structural similarity across the board, even compared with GPT-3.5/4.\footnote{Our code is available at \url{https://github.com/launchnlp/NoT}.} 
\end{abstract}

%% file: 1_introduction.tex
\section{Introduction}

\begin{figure}[t]
    \centering
    \includegraphics[width=0.36\textwidth]{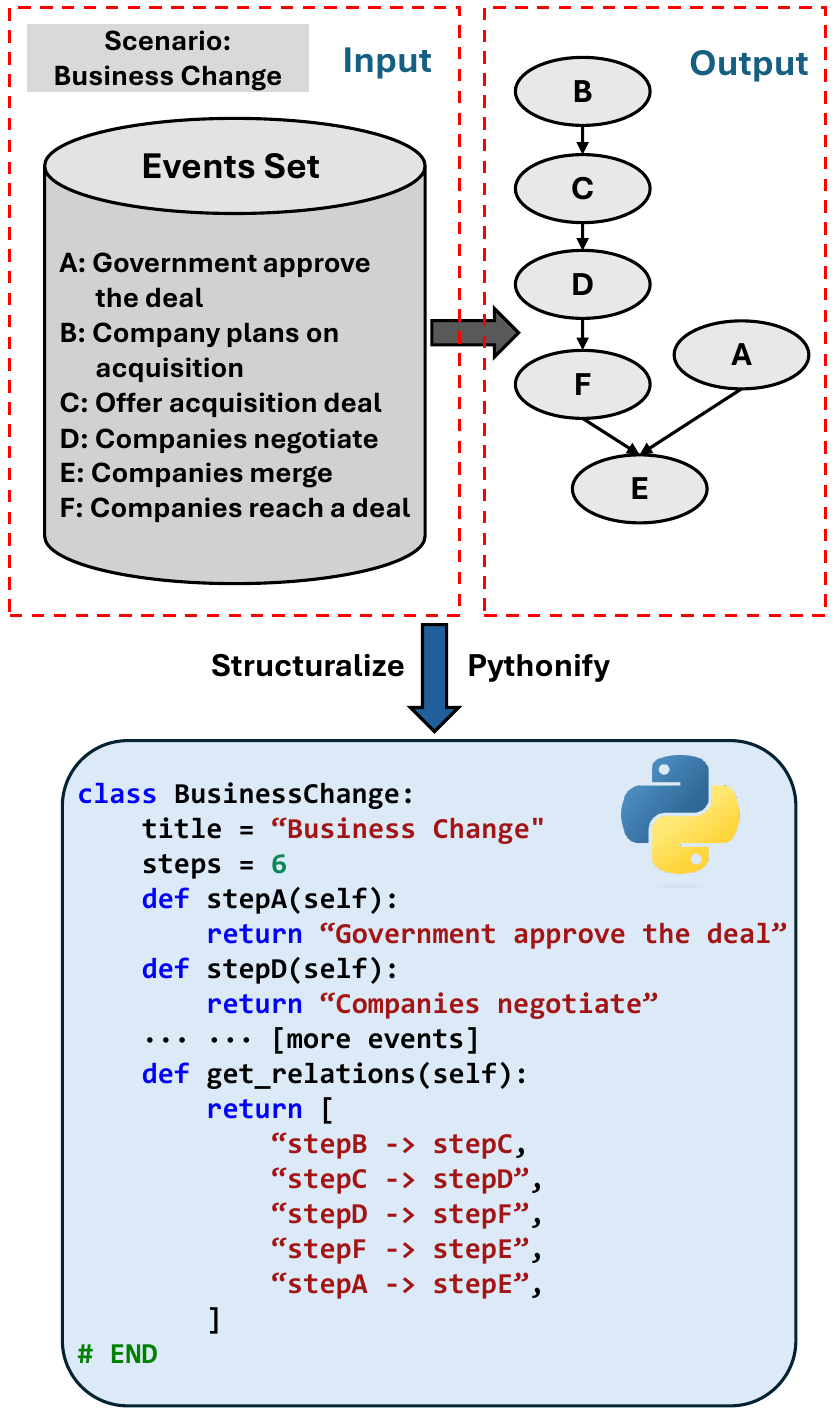}
    \caption{Task overview of \textbf{temporal graph generation (TGG)}, where the input is a goal and a set of unordered events. In this work, to better unleash the pre-training power of LLMs trained with a mixture of text and code, we cast TGG as a code completion task.
    }
    \label{fig:task}
    \vspace{-6mm}
\end{figure}

Temporal reasoning is essential for humans to perceive the world, understand daily communications, and interpret the temporal aspects of experiences~\citep{journals/cacm/Allen83, journals/jacm/NebelB95}. 
The recent advent of large language models (LLMs) has garnered substantial attention to their impressive performance in various reasoning tasks, such as arithmetic reasoning \citep{gsm8k, zhong2024achieving} and commonsense reasoning \citep{talmor-etal-2019-commonsenseqa, palm2}.
Nonetheless, few LLMs exist to handle temporal reasoning well \citep{time_benchmark1, time_benchmark2, chan-etal-2024-exploring}, due to the task's inherent complexity, mingled with implicit logical inference and the necessity for profound world knowledge. 

To gain deeper insights, the research community mainly focuses on two ends along the spectrum: either a simple relation extraction task that orders a pair of events \citep{ uzzaman-etal-2013-semeval, yuan-etal-2023-zero}, or a perplexing 
commonsense understanding task demanding multifaceted reasoning skills beyond the mere temporal aspect \citep{journals/corr/abs-2308-00002, tan-etal-2023-towards, Xiong-LTR}. Worse still, the former is limited to a \textit{local} scope spanning two adjacent sentences only and fails to account for the significance of \textit{global} temporal relations, leading to overly optimistic results \citep{restructured, time_benchmark1}. 
Therefore, neither setup provides a clear understanding of LLMs' true temporal reasoning abilities. 

In this work, we aim to unveil the \textbf{inherent, global temporal reasoning capabilities of LLMs}, evaluating them in isolation \textit{free from confounding factors}, and addressing the limitations of previous studies which only focused on local contexts. 
We first introduce a task of \textbf{temporal graph generation (TGG; \Cref{fig:task})}: Given a high-level goal\footnote{We use \textit{goal} and \textit{scenario} interchangeably.} $\mathcal{T}$
(e.g., business change) and a set of events $\mathcal{V}$, the objective is to produce a temporal graph $\mathcal{G(V, E)}$ where a directed edge in $\mathcal{E}$ reveals the temporal order between events. 
Though this specific notion of TGG is new, many of its applications are not. 
In this work, we specifically study TGG in order to evaluate and improve the temporal reasoning capability, since TGG is deemed a major bottleneck when LLMs perform temporal reasoning. With TGG, we put forth the first research question.

\noindent\textbf{RQ1: What is the temporal reasoning capability of popular LLMs?} 
Prior work~\citep{time_benchmark1, time_benchmark2} shows a huge gap between AI systems and human performance on various temporal understanding tasks. Additionally, there is a notable performance disparity between proprietary LLMs (e.g., GPT-4) and open-weights LLMs, particularly those with fewer than 10 billion parameters (henceforth, small LLMs). 
Our study on temporal reasoning reveals a similar trend and identifies the existence of both gaps, as demonstrated in \Cref{tbl:main_results}. This further highlights the importance of an in-depth investigation of TGG, since the performance of downstream tasks (e.g., temporal commonsense understanding) is positively correlated with the inherent, global temporal reasoning capability. 
Observing the model deficiencies, we are motivated to \textit{fill the gap between open-weights, small LLMs and proprietary large models}. This is due to the fact that open-weights LLMs are generally more accessible, reproducible, and cost-effective to use \citep{chatgpt-oneyear, ethicalGPT}. In pursuit of this goal, we present the second research question.

\noindent\textbf{RQ2: With a budget constraint (e.g., not allowing further training), how can small LLMs catch up with large models like GPT-3.5/4?} 
Given the constraint that no training will be used, we propose \methodFull (\method), a special prompting technique tailored for temporal reasoning. This method capitalizes on the recent success of the Chain-of-Thought (CoT) technique \citep{Wei0SBIXCLZ22, KojimaGRMI22}, found effective in solving complex reasoning tasks. To approach TGG, \method produces a final temporal graph via first generating a \textit{temporally grounded narrative}\footnote{In our context, ``temporally grounded'' refers to events being organized and presented in a way that accurately reflects their temporal sequence or timeline.} then sorting the input events topologically in reference to the recounted narrative. Inspired by \citet{madaan-etal-2022-language, PoT, PAL}, \method also features structural representations by converting the input-output mapping to a Python class, and instructing the generation in code space. 
We further improve \method by introducing high-quality reference narratives as part of few-shot demonstrations.

Extensive experiments across three evaluation benchmarks of diverse genres reveal six interesting findings: 1) small LLMs \textit{struggle with temporal reasoning} {even with few-shot examples}; 2) \textit{CoT is also ineffective at temporal reasoning}, in line with existing finding \citep{time_benchmark2}; 3) \textit{GPT-4 sometimes falls off the throne due to alignment}, when answering sensitive queries;
4) \method is a powerful tool to assist small LLMs to catch up with or even \textit{surpass GPT-3.5}, and presents strong compatibility with various base LLMs;  5) \textit{the temporally grounded narratives are significant in improving LLMs' temporal reasoning process}; 6) \textit{AI systems are far from mastering temporal reasoning}, trailing the human baseline by 30 F1 points. 
 
We also analyze the impact of shot numbers and perform a holistic evaluation of reference narratives in few-shot examples. 5-shot is found to be the sweet spot for temporal reasoning, after which the performance plateaus, {likely due to long-context challenge}. We identify three key characteristics of reference narratives for them to avail small LLMs most: conciseness, simplicity, and factuality.

%% file: 2_related_word.tex
\section{Related Work}
\subsection{Temporal Reasoning}

This work is deeply rooted in a long-standing yet still challenging NLP domain---temporal reasoning \citep{journals/cacm/Allen83, journals/jacm/NebelB95}, which involves extraction, representation and reasoning with time and events \citep{Sanampudi2010TemporalRI}.
Depending on the cognitive complexity, temporal reasoning in NLP is studied at three levels: 
temporal expression detection, temporal relation extraction, and temporal graph generation. The simplest \textbf{temporal expression detection} task is to identify phrases in the text that convey temporal information \citep{Setzer01, mani-etal-2001-guidelines, PustejovskyCISGSKR03}, commonly known as TimeX. 
Further, under-specified TimeX is typically converted to explicit expressions (e.g., ``summer 2024'') through a process called {time expression normalization} \citep{verhagen-etal-2010-semeval}. 

Explicit TimeX is often absent in text, and events usually carry implicit temporal information. To bridge the gap, TempEval~\citep{Verhagen2009TheTC, uzzaman-etal-2013-semeval} is curated to support the study of \textbf{temporal relation extraction},  which aims to detect the temporal relation between two \textit{events} in a document.  The most common benchmarks, TB-dense \citep{chambers-etal-2014-dense} and MATRES \citep{ning-etal-2018-multi}, have witnessed the technique evolution from LSTM \citep{dligach-etal-2017-neural} and GNN-augmented BERT \citep{mathur-etal-2021-timers, wang-etal-2022-dct}, to LLMs prompting \citep{yuan-etal-2023-zero}. Yet, these benchmarks are limited by their \textit{locality assumption}, where only pairs of events within a two-sentence window are annotated. Even in this simplified scenario of temporal relation extraction, ChatGPT perform poorly, trailing supervised systems by over 30\% \citep{chan-etal-2024-exploring}.

The most challenging task, \textbf{contextualized temporal graph extraction}, is defined as, given a document, generating a corresponding event-level temporal graph \citep{uzzaman-etal-2013-semeval, madaan-yang-2021-neural}. This task addresses the limitation of locality by priming models to comprehend the entire article and infer relationships even between distant events. Yet, this area is largely under-investigated, partly due to the scarcity of available datasets.
 A similar task is \textbf{script learning} \citep{regneri-etal-2010-learning, modi-etal-2016-inscript, sakaguchi-etal-2021-proscript-partially},
 which targets inducing a stereotypical progression of \textit{complex} events \citep{script_def}, represented as a temporal graph of more \textit{atomic} events. This task is usually approached by first extracting information snippets from a given document to build an instance graph, and then expanding the graph to generate a schematic graph using GNN \citep{li-etal-2021-future, jin-etal-2022-event} or LLM prompting \citep{dror-etal-2023-zero}. 
 Given the remarkable similarities between these two tasks, we instead study a temporal reasoning task formulation that is \textit{fundamental} to both, i.e., \textbf{temporal graph generation}. It differs from prior work in at least two dimensions: (1) a limited-context setting, where only abstract event descriptions are available, and (2) only a few training samples at hand, rendering fine-tuning techniques inapplicable. 
 This motivates a \textit{training-free assessment} of LLMs' \textit{inherent, global} temporal reasoning capability. 

\begin{figure*}[t]
    \centering
    \includegraphics[width=0.9\textwidth]{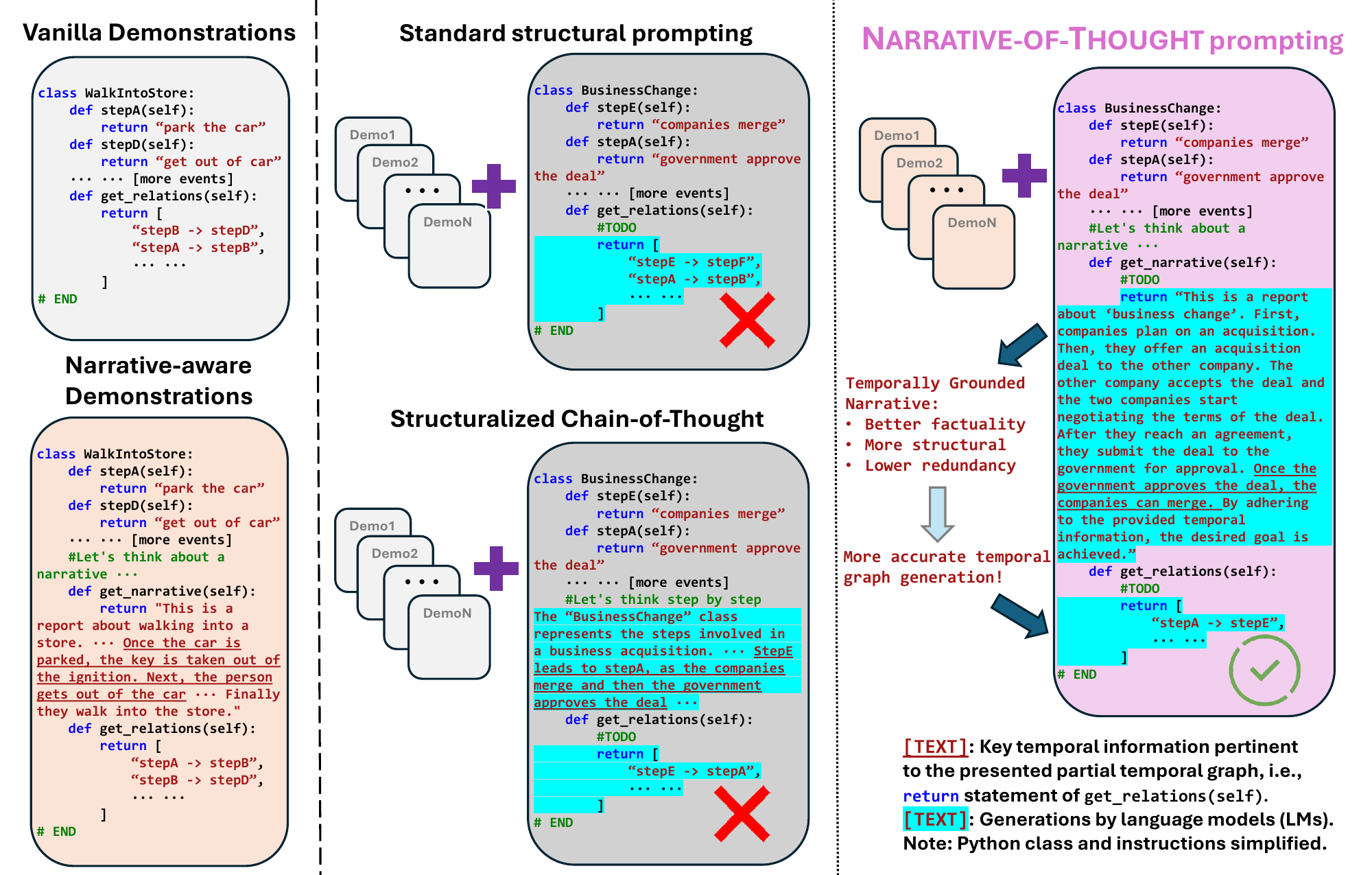}
    \vspace{-2mm}
    \caption{Overview of \methodFull (\method), a prompting technique tailored for temporal reasoning. \method improves the temporal graph by \textit{recounting} a temporally grounded narrative. Also shown are comparisons with existing methods.  Full example is in \Cref{fig:full_example_gensort} and \method output is in \Cref{fig:output_gensort}.
    }
    \label{fig:architecture}
    \vspace{-4mm}
\end{figure*}

\subsection{Chain-of-Thought and its Variants}
\label{sec:CoT}

Despite the strong problem-solving capability in the general domain \citep{WeiTBRZBYBZMCHVLDF22}, LLMs struggle to address more complex reasoning tasks, such as commonsense understanding and arithmetic reasoning \citep{patel-etal-2021-nlp, TalmorYBBGCB21, huang-chang-2023-towards}. \citet{Wei0SBIXCLZ22} first introduce the concept \textit{Chain-of-Thought (CoT)} by decomposing multi-step problems into intermediate steps. \citet{KojimaGRMI22} further adds a phrase \textit{``Let's think step by step''} to perform zero-shot CoT. These studies underpin the CoT technique in enhancing LLMs' capability for complex reasoning. 

Down the line, sophisticated prompting schemes are devised through \textit{structuralization}.  One approach is to extend the linear chain structure to Tree-of-Thoughts \citep{yao2023tree} and Graph-of-Thoughts \citep{besta2024got}, enabling expanded exploration space. The huge search space, however, results in a computational resource dilemma. On top of that, leveraging the deterministic execution to narrow the discrepancy between reasoning and final answer,  PoT \citep{PoT}, PAL \citep{PAL} and Faithful CoT \citep{lyu-etal-2023-faithful} introduce programming languages to describe the reasoning process structurally.
These methods are designed exclusively for solving mathematical reasoning and symbolic reasoning, where the reasoning process and computation can be decoupled. In contrast, for temporal reasoning, the reasoning process and the temporal sorting step are intrinsically interleaved. In fact, \citet{time_benchmark2} has attempted to apply CoT but proved unsuccessful.

Moreover, existing methods are mostly applied to generate intermediate rationales for \textit{simple, atomic outputs}, usually in the format of multi-choice options \citep{mihaylov-etal-2018-suit, talmor-etal-2019-commonsenseqa, logicalQA}, a number \citep{gsm8k, math}, or yes/no options~\citep{commonsenseqa2, WeiTBRZBYBZMCHVLDF22}. Our work draws a clear distinction where our focus is on \textbf{structural output generation}, augmented with producing a rationale in the form of a compelling and pertinent narrative.\footnote{The significance of narrative in shaping human decision-making is well-studied \citep{piper-etal-2021-narrative, emelin-etal-2021-moral, zhang-etal-2024-moka}; we hypothesize machines are similarly influenced.}

%% file: 3_method.tex
\section{Method: \methodFull}

\Cref{fig:architecture} provides an overview of the proposed \methodFull (\method) method, and draws a comparison against common prompting techniques. Overall, given a scenario and a set of events, \method first converts the input into a Python class, then guides LLMs to produce a temporally grounded narrative by arranging events in the correct temporal order, leveraging LLMs' intrinsic temporal knowledge. Based on the \textit{recounted} temporal relations articulated in the narrative, LLMs are instructed to sort events into a temporal graph. This section will discuss major components in detail: (1) structural representation, (2) \method prompting template, and (3) narrative-aware demonstrations.

\paragraph{Structural Representation.} Following prior work~\cite{madaan-etal-2022-language, PoT, PAL}, we cast temporal reasoning as a code completion task. This design decision is motivated by the unordered nature of both event sets and temporal relation sets, making a structural representation the optimal choice. \citet{wang-etal-2023-code4struct} also shows that combining structural event representations with LLMs trained with a mixture of text and code can unleash the full pretraining power. 
We extend this framing to handle cross-event structures. Specifically, a temporal graph is commonly presented in DOT format \citep{madaan-yang-2021-neural, sakaguchi-etal-2021-proscript-partially}, the appearance of which lends itself naturally to the usage of coding format.
Furthermore, code execution follows a clear, step-by-step logical flow, mirroring the process of reasoning. 
Bringing these aspects together results in an alignment between temporal graphs and code structure, facilitating the temporal reasoning process. {Our further study on this phenomenon also reveals a strong positive correlation between coding capabilities and temporal reasoning, as documented in \Cref{sec:correlation_analysis} .}

Concretely, each scenario is represented as a Python class. Each class encapsulates events as functions, where the function name is in the form of ``step[\texttt{A-Z}]'' such as ``step\texttt{X}'',  
and the function body indicates the event description. The temporal graph is represented as a collection of pairwise temporal relations, enclosed within the return statement of ``get\_relation()'' function, marked by ``TODO'' for LLMs to implement.

\paragraph{\methodFull (\method).}

At inference time, \method first prompts LLMs to produce a temporally grounded narrative using \textit{Narrative Prompt}. Drawing on the generated narrative, LLMs proceed and complete generation in response to \textit{Temporal Graph Prompt}. The entire generation process is in an end-to-end manner, ensuring that LLMs explicitly leverage the temporal relations articulated in the narrative to assist the generation of the final temporal graph. We provide a complete example in \Cref{appx:outputs}.

\begin{tcolorbox}[height=2.5cm, title=Narrative Prompt, fontupper=\small]
\# Let's think of a narrative to link aforementioned events in the correct temporal order.   

def get\_narrative(self):

\# TODO   
\end{tcolorbox}

 \vspace{-2mm}

\begin{tcolorbox}[height=2.1cm, title=Temporal Graph Prompt, fontupper=\small]
def get\_relations(self):

\# TODO

\# END
\end{tcolorbox}
 \vspace{-2mm}

Overall, \method narrows the gap between pre-training and inference by allowing the LLM to unfold the narrative knowledge seen during pre-training. 
Concretely, our approach leverages LLMs' inherent strengths in \textit{generating} and \textit{comprehending} text for narrative and temporal graph generation, respectively. In contrast, directly mapping abstract events to a temporal graph is less effective, as such examples are rarely encountered during pre-training. 
Practically, generated narratives create imagined experiences to navigate, and reify implicit timelines, assisting reasoning over a series of events even without explicit timestamps provided in the text,  which are crucial for tasks requiring temporal reasoning.  
By reading the \textit{recounted} narrative, it becomes easier for the LLMs to construct an implicit timeline to guide event sorting, significantly reducing the reasoning complexity compared to generating temporal graphs from scratch (i.e., using abstract events alone).

Our \method draws a clear distinction from the CoT prompting and its variants in four aspects. First, for CoT, a final answer cannot be easily extracted unless a post-hoc script is designed \citep{KojimaGRMI22, self-consistency, stepback}, which can be sometimes error-prone, while the output of \method is easy to obtain by parsing the \texttt{get\_relations()} function. 
Second, \method produces final outputs in the structural space, while existing methods solely produce \textit{simple, atomic outputs} 
as discussed in \Cref{sec:CoT}.
Third, \method produces final temporal graphs cost-effectively without external tools in an end-to-end fashion, unlike pipeline approaches which face error propagation and over-sampling issues \citep{dror-etal-2023-zero}. 
Lastly, the generated rationales by CoTs are not necessarily grounded in real-world experience.  
In contrast, generated narratives by \method are steered to be more \textit{temporally grounded}, creating an imagined experience for LLMs to navigate, which is proved effective.

\paragraph{Narrative-aware Demonstrations.}
Existing studies~\citep{gpt3, WeiTBRZBYBZMCHVLDF22} have demonstrated that in-context demonstrations play a critical role in guiding LLMs to produce meaningful outputs. \method is no exception, as  \Cref{tbl:main_results} reveals that even GPT-3.5 struggles with temporal reasoning in a zero-shot setting.
Thus, few-shot examples are provided by default. 
For \method to succeed, high-quality and relevant rehearsed narratives, termed \textit{reference narratives}, need to be created and embedded in these demonstrations.

Capitalizing on the recent success of using LLMs to generate demonstrations \citep{ generate-then-read, self-demonstration}, we prompt GPT-3.5/4 to produce reference narratives. Concretely, for each demonstration, abstracted as $\mathcal{G(V, E)}$, we feed both $\mathcal{V}$ and $\mathcal{E}$ into GPT-3.5/4, using our designed reference narrative generation templates, dubbed \textit{meta prompts}. In total, we create 4 types of meta prompts covering diverse genres like news and children's stories. 
Additionally, when feeding $\mathcal{G(V, E)}$ into GPT-3.5/4, we use two \textit{input formats} to define a Python class (\textit{alphabetical} like ``step\texttt{X}'' in \Cref{fig:meta_alphabetical} vs. descriptive like ``pushPedal'' in \Cref{fig:meta_descriptive}). We later evaluate the usefulness of each meta prompt in \Cref{sec:ablation_study}. Details of meta prompts are documented in \Cref{appx:meta_prompt}.

%% file: 4_experiment.tex
\section{Experiment}

\input{tables/main_results}

In this work, we focus on \textbf{Temporal Graph Generation (TGG)}, an essential task of temporal reasoning. Here, we discuss datasets, experimental setup, baselines, and evaluation metrics. We provide additional implementation details in \Cref{appx:implementation}.

\subsection{Dataset}
In line with the literature, we use \textbf{ProScript} \citep{sakaguchi-etal-2021-proscript-partially} as the major benchmark,  where a temporal script is represented as a directed acyclic graph, which were collected from a diverse range of sources including ROCStories \citep{mostafazadeh-etal-2016-corpus}, Descript \citep{wanzare-etal-2016-crowdsourced}, and Virtual home \citep{PuigRBLWF018}. 
We also adopt two other datasets to enrich the evaluated genres and domains, and make necessary changes for the TGG task: 1) \textbf{Schema-11} evaluation set \citep{dror-etal-2023-zero}, which contains human-curated event schemas for 11 newsworthy topics, such as \textit{armed robbery} and \textit{business change}; 
and 2) \textbf{WikiHow Script} corpus \citep{lyu-etal-2021-goal}, a collection of multilingual how-to articles depicting necessary steps performed in sequence to achieve a high-level goal, covering a wide range of daily activities.
Dataset statistics are included in \Cref{tbl:data_stats}, and we provide detailed dataset processing steps in \Cref{appx:dataset_processing}.

\subsection{Setup}
\label{sec:exp_setting}
As our goal is to study the capability and generalizability of existing LLMs, and our \method without any fine-tuning, we assume no access to large-scale training sets except for few-shot demonstrations. Therefore, all experiments are conducted in a 5-shot setting. We provide analysis on the impact of the shots numbers in \Cref{sec:ablation_study}. We consider three base models to spotlight the compatibility and versatility of \method. We include very recent, strong LLMs, showing promising results on various reasoning tasks and code completion tasks, \textsc{Mistral}-7B \citep{mistral}, \textsc{Gemma}-7B \citep{gemma}, and \textsc{Llama3}-8B \citep{llama3modelcard}. For all base models, we use their instruction-fine-tuned versions for experiments.

\input{tables/data_stats}

Shown in \Cref{fig:architecture}, we represent the event set as a suite of Python methods, by serializing the unordered event set. For each scenario, we randomly shuffle the input Python methods three times, and apply models to each shuffle with greedy decoding at inference. For \method, we use \textit{Simple Report}-style narratives by GPT-4, {which are generated by following instructions to produce concise reports based on provided event descriptions and relations (see style details in \Cref{tbl:meta_prompt_type}).}

\subsection{Baselines}
To showcase the effectiveness of \method, for each base model 
we compare with standard structural prompting and structuralized chain-of-thought prompting
(\Cref{fig:architecture}). We also remove reference narratives in demonstrations to highlight the importance of narrative-aware few-shot demonstrations, and conduct a holistic evaluation of reference narratives in \Cref{sec:ablation_study}. We include a random baseline, where events are naively connected to form a \textit{linear} temporal chain based on the order they appear in the input. We also experiment with two strong proprietary models, GPT-3.5\footnote{\url{https://chat.openai.com/}; \texttt{gpt-35-turbo-16k-0613}, training data up to Sept. 2021.} and GPT-4 \citep{gpt-4}\footnote{\texttt{gpt-4-turbo-0125-preview},  data up to Dec. 2023.} to help gauge the gap between AI systems and human-level performance.

\subsection{Evaluation Metrics}
We denote the ground-truth and generated temporal graphs as $\mathcal{G(V,E)}$ and ${\hat{\mathcal{G}}(\mathcal{V},\hat{\mathcal{E}})}$, respectively. we compare both semantic and structural similarities between  $\mathcal{G}$ and $\hat{\mathcal{G}}$, following prior work~\citep{sakaguchi-etal-2021-proscript-partially, madaan-etal-2022-language}. To evaluate semantic similarity, we report \textit{precision (P)} and \textit{recall (R)}, defined as below, as well as \textit{F1}.

\vspace{-2mm}
\small
\begin{equation*}
  \mathrm{Precision} = \frac{|\mathcal{E} \cap \hat{\mathcal{E}}|}{|\hat{\mathcal{E}}|}  \;\;\; \mathrm{Recall} = \frac{|\mathcal{E} \cap \hat{\mathcal{E}}|}{|\mathcal{E}|} 
\end{equation*}

\normalsize
\noindent To assess structural similarities, we consider:
\begin{itemize}[leftmargin=2em,itemsep=0em,topsep=1pt,parsep=1pt,partopsep=1pt]
    \item \textit{Graph Edit Distance} \citep[\textit{GED};][]{Abu-AishehRRM15} calculates the minimum number of edits (node/edge removal/additions) to transform $\hat{\mathcal{G}}$ to a graph isomorphic to $\mathcal{G}$.
    \item \textit{Graph Statistics}: fraction of the number of edges between $\hat{\mathcal{G}}$ and $\mathcal{G}$ ($\frac{|\hat{\mathcal{E}}|}{|\mathcal{E}|}$); the number of connected components in $\hat{\mathcal{G}}$, denoted as $k(\mathcal{G})$. The  goal is to bring both statistics closer to 1, additionally ensuring $k(\mathcal{G})$ is at least 1.
\end{itemize}

\noindent We further calculate \textit{Pair-wise Consistency} between $\hat{\mathcal{G}}_i$ and $\hat{\mathcal{G}}_j$, where we compare generated graphs, based on two randomly shuffled inputs, and compute the proportion of common temporal links produced in both graphs, i.e.,  $\frac{|\hat{\mathcal{E}}_i \cap \hat{\mathcal{E}}_j|}{|\hat{\mathcal{E}}_i \cup \hat{\mathcal{E}}_j|}$. 

%% file: tables/main_results.tex
\begin{table*}[t]
\centering
\resizebox{0.94\linewidth}{!}{%
\begin{tabular}{lrrrrrrrrrrrrrr}
\toprule
\multicolumn{1}{c}{\multirow{2}{*}{Method}} & \multicolumn{4}{c}{Proscript}                                 & \multicolumn{4}{c}{Schema-11}                                       & \multicolumn{4}{c}{WikiHow Script}                            & \multicolumn{2}{c}{Avg.}                    \\ 
 \cmidrule(lr){2-5}
  \cmidrule(lr){6-9}
   \cmidrule(lr){10-13}
    \cmidrule(lr){14-15}
\multicolumn{1}{c}{}                        & F1$\uparrow$   & GED$\downarrow$  & $k(\mathcal{G})$ & Cons.$\uparrow$  & F1$\uparrow$   & GED$\downarrow$  & $k(\mathcal{G})$ & Cons.$\uparrow$  & F1$\uparrow$   & GED$\downarrow$  & $k(\mathcal{G})$ & Cons.$\uparrow$         & F1$\uparrow$                   & GED$\downarrow$                  \\ \midrule
\multicolumn{15}{c}{Baselines}     \\ \midrule
Random                                      & 14.0          & 1.47          & 1.00          & 7.8           & 19.4          & 3.91                & 1.00          & 7.8           & 14.2          & 0.06          & 1.00          & 8.8           & 15.9                 & 1.81                 \\
GPT-3.5 (0-shot)*                            & 18.4          & 2.25          & 1.06          & 38.6          & 30.1          & 4.48                & 1.27          & 30.2          & 17.2          & 2.80          & 1.11          & 40.8          & 21.9                 & 3.18                 \\
GPT-3.5                                     & 43.4          & 1.71          & 1.07          & 38.8          & 62.8          & 3.30                & 1.36          & 50.2          & 31.0          & 1.58          & 1.10          & 35.4          & 45.7                 & 2.20                 \\
GPT-4                                       & 63.9          & 1.64          & 1.02          & 61.4          & 44.1          & 7.97                & 0.64          & 46.3          & 43.0          & 1.71          & 1.04          & 48.5          & 50.3                 & 3.77                 \\ \midrule
\multicolumn{15}{c}{\textsc{Gemma-7B} \citep{gemma}}  \\ \midrule
Standard Prompting                          & 19.7          & \textbf{2.35} & 1.02          & \textbf{20.4} & 27.8          & 5.03                & \textbf{1.03} & 18.3          & 17.5          & \textbf{2.88} & \textbf{0.96} & \textbf{17.3} & 21.7                 & \textbf{3.42}        \\
Chain-of-Thought                            & 20.0          & \textbf{2.35} & 1.01          & 20.0          & 26.4          & 5.03                & \textbf{1.03} & 14.9          & 13.6          & 5.91          & 0.73          & 11.5          & 20.0                 & 4.43                 \\ \hdashline[5pt/4pt]
\method (no reference)            & 20.0          & 2.47          & \purple{\textbf{1.00}} & 17.3          & \red{27.9}          & \red{\textbf{4.78}}       & 1.09          & 18.1          & 15.2          & 5.03          & 0.81          & 13.9          & 21.0                 & 4.09                 \\
\method (alphabetical meta)              & \red{\textbf{21.8}} & 2.48          & \purple{\textbf{1.00}} & 18.3          & \red{\textbf{36.0}} & \red{4.84}                & 1.06          & \red{19.7}          & \red{\textbf{17.9}} & 2.95          & \textbf{0.96} & 16.9          & \red{\textbf{25.2}}        & \textbf{3.42}        \\
\method (descriptive meta)          & \red{21.3}          & 2.60          & 0.99          & 17.8          & \red{34.8}          & \red{5.00}                & 1.06          & \red{\textbf{20.8}} & \red{\textbf{17.9}} & \textbf{2.88} & 0.95          & 16.8          & \red{24.7}                 & 3.49                 \\ \midrule
\multicolumn{15}{c}{\textsc{Mistral-7B} \citep{mistral}}  \\ \midrule
Standard Prompting                          & 30.7          & 2.16          & 1.05          & 22.3          & 35.3          & 4.55                & 1.12          & 29.1          & \textbf{22.5} & \textbf{2.09} & 1.11          & \textbf{18.9} & 29.5                 & 2.93                 \\
Chain-of-Thought                            & 29.8          & 2.66          & \textbf{1.02} & 22.1          & 35.2          & 5.33                & 0.94          & 30.5          & 20.5          & 2.59          & 1.10          & 17.4          & 28.5                 & 3.53                 \\ \hdashline[5pt/4pt]
\method (no reference)            & \red{32.5}          & 3.04          & 0.95          & 19.4          & \red{42.3}          & 5.27                & \purple{\textbf{1.00}} & \red{27.6}          & 21.8          & 3.33          & 0.98 & 15.4          & \red{32.2}                 & 3.88                 \\
\method (alphabetical meta)              & \red{35.2}          & \red{\textbf{2.11}} & \textbf{1.02} & \red{22.4}          & \red{50.9}          & \red{4.30}                & \purple{1.03}          & \red{\textbf{36.1}} & 21.7          & 2.49          & \red{\textbf{1.04}}          & 14.8          & \red{35.9}                 & 2.97                 \\
\method (descriptive meta)          & \red{\textbf{35.4}} & \red{2.14}          & \textbf{1.02} & \red{\textbf{23.0}} & \red{\textbf{52.7}} & \red{\textbf{3.90}}       & \purple{1.06}          & \red{32.5}          & 22.1          & 2.53          & \red{\textbf{1.04}}          & 15.1          & \red{\textbf{36.7}}        & \red{\textbf{2.86}}        \\ \midrule
\multicolumn{15}{c}{\textsc{Llama3-8B} \citep{llama3modelcard}} \\ \midrule
Standard Prompting                          & 25.1          & 2.39          & 1.18          & 19.9          & 28.3          & 4.42                & 1.24          & 19.9          & 20.6          & 1.17          & 1.07          & 21.2          & 24.7                 & 2.66                 \\
Chain-of-Thought                            & 30.1          & 2.06          & \textbf{1.00} & 23.3          & 37.3          & 5.79                & 0.85          & 23.5          & 22.6          & \textbf{0.99} & \textbf{1.02}          & \textbf{24.3} & 30.0                 & 2.95                 \\ \hdashline[5pt/4pt]
\method (no reference)            & \red{35.5}          & \red{1.88}          & \textbf{1.00} & \red{25.3}          & \red{52.6}          &  \purple{\textbf{3.18}} & \purple{1.12}          & \red{35.0}          & \red{25.4}          & \blue{\textbf{0.99}} & \blue{\textbf{1.02}}          & 20.9          & \red{37.8}                 & \purple{\textbf{2.02}}        \\
\method (alphabetical meta)              & \red{\textbf{39.5}} & \red{1.87}          & \blue{1.01}          & \red{\textbf{28.8}} & \red{59.0}          & \red{3.72}               & \purple{1.12}          & \red{39.1}          & \red{26.3}          & \blue{1.01}          & \blue{1.03}          & 22.5          & \red{41.6}                 & \blue{2.20}                 \\
\method (descriptive meta)          & \red{38.7}          & \red{\textbf{1.86}} & \blue{1.01}          & \red{28.4}          & \red{\textbf{61.5}} & \red{3.57}                & \purple{\textbf{1.09}} & \red{\textbf{45.6}} & \red{\textbf{26.5}} & \blue{1.04}          & \blue{1.03}          & 22.3          & \red{\textbf{42.2}}        & \purple{2.16}                \\ \bottomrule
\end{tabular}
}
\vspace{-2mm}
\caption{
Main results of base LLMs and strong baselines on TGG evaluation benchmarks (average of 3 runs). For each base model, best results are \textbf{bold}, and \method's variants better than both Standard Prompting and CoT are \red{highlighted}. \method results that outperform 5-shot GPT-3.5 and GPT-4 are in \blue{blue}. Results that meet both criteria are in \purple{purple}.
{On average, \method boosts F1 metric over its base model by 16\% to 71\%, and sometimes improves the GED metric. \method-augmented \textsc{Llama3-8B} achieves best overall F1 (63.5 F1 by 3-shot variant; Figure~\ref{fig:shots}) and GED results on Schema-11. Also, it only trails GPT-3.5 and GPT-4 by 8\% and 14\% on average, while yielding a lower average GED.} 
Full results in Table~\ref{tbl:full_results}. *By default, 5-shot examples are provided, unless otherwise noted.
}
\label{tbl:main_results}
\end{table*}

%% file: tables/data_stats.tex
\begin{table*}[t]
\centering
\resizebox{0.90\linewidth}{!}{%
\begin{tabular}{lrrrrrrr} \toprule
\multicolumn{1}{l}{} & \#scenarios & \#events & Max \#events & \#temporal links & Event length & \%Non-linear & Domain       \\ \midrule
ProScrpt (\citeauthor{sakaguchi-etal-2021-proscript-partially})             & 2,077         & 7.46      & 9                      & 6.95              & 4.64                      & 39\%               & Daily\\
Schema-11 (\citeauthor{dror-etal-2023-zero})           & 11           & 7.91      & 11                     & 7.18              & 3.48                      & 27\%               & News         \\
WikiHow Script (\citeauthor{lyu-etal-2021-goal})      & 2,991          & 8.37      & 20                     & 7.37              & 9.63                      & 0\%                & Daily \\ \bottomrule
\end{tabular}
}
\vspace{-2mm}
\caption{Basic statistics of evaluation datasets. Max \#events indicate the maximum number of events for a scenario.  Event length is defined as the number of words in the event description. \%Non-linear tells the proportion of temporal graphs that contain at least one branch. Two domains are considered, \textit{Daily} activity and \textit{News} journalism.}
\label{tbl:data_stats}
\end{table*}

%% file: 5_results_analysis.tex
\section{Results and Analyses}

\subsection{Main Results}
\label{sec:main_results}
Major results are included in \Cref{tbl:main_results}, and the full results (across all 7 metrics) can be found in \Cref{tbl:full_results}. Below are our major findings.

1)  \textit{With the few-shot setup, small LLMs are dramatically underperforming, reaching barely 50\% of GPT-4's capabilities.} 
The three base models, whether using standard prompting or CoT, consistently under-perform GPT-4 and attain 40\% to 60\% of its average F1 scores. Among them, \textsc{Mistral-7B} achieves the highest F1 scores, while \textsc{Llama3-8B} produces temporal graphs most similar to the ground truth, as measured by GED.

2) \textit{Unlike many other reasoning tasks, CoT does not always work for temporal reasoning and sometimes degrades performance.}  Unlike mathematical or logical reasoning \citep{Wei0SBIXCLZ22}, CoT prompting does not necessarily enhance model performance on temporal reasoning tasks. Across all three base models, there is a notable degradation in F1 and GED scores with CoT, except for \textsc{Llama3}'s F1 scores. This is not TGG-specific, but rather a common pattern across various temporal understanding tasks \citep{time_benchmark2}, highlighting the need for specialized approaches to temporal reasoning.
Outputs by CoT are included in \Cref{fig:output_CoT}.
 
3) \textit{GPT-4 is not always the champion, owing to the added safety layer.} 
GPT-4 implements safety measures through human-preference alignment \citep{gpt-4}, which enhances model safety by prompting more cautious responses, potentially leading to performance drop \citep{alignment-1, alignment-2}. Especially on \textbf{Schema-11},  GPT-4 refrains from providing answers to sensitive scenarios like ``bombing attacks'',\footnote{In our experiments, we disabled content filtering.} and thus fails to produce a valid temporal graph. 

4) \textit{With \method, small LLMs can perform comparably to GPT-3.5, or even take the lead.} When equipped with \method, the overall semantic correctness (F1) and structural similarity (GED) of the generated temporal graphs are significantly enhanced, regardless of which base LLM is used.
The average improvement of F1 over naively prompting the base model is between 16\% to 71\%. As the power of the base LLM grows, \method demonstrates greater consistency in its outputs.
Notably, with \textsc{Llama3-8B}, the strongest base LLM, \method achieves an F1 score that is comparable to GPT-3.5 (42.2 vs. 45.7), and even outperforms GPT-3.5/4 on GED. These  results demonstrate the potential of applying \method in a wide range of temporal understanding tasks in future research. 

5) \textit{Recounting temporally grounded narrative is a prerequisite for LLMs to generate temporal graphs accurately.} 
Without high-quality reference narratives,
LLMs struggle to generate temporally grounded narratives, leading to a detrimental impact on \method-augmented \textsc{Gemma-7B} (e.g., a 0.7 F1 drop and a 0.67 GED increase).

\input{tables/FT_results}
6) \textit{
LLMs, including the powerful GPT-4, lag far behind human-level performance in temporal reasoning.}
The SOTA F1 score (by GPT-4) on ProScript is 63.9, 
whereas the human baseline F1 is 89.3 \citep{sakaguchi-etal-2021-proscript-partially}.
While \method has notably narrowed the gap between small and large LLMs, AI models have not mastered temporal reasoning yet, and further research efforts are needed for LLMs to match human performance.

\paragraph{Comparison with fine-tuned LLMs.} To evaluate the performance gap between the \method prompting technique and the computational-intense fine-tuning (FT) approach, we conduct a side experiment on the ProScript dataset. Specifically, each instruction-tuned base LLM is fine-tuned on the ProScript training set, utilizing LoRA \citep{lora} and mixed-precision training. We follow the same setting as in \Cref{sec:exp_setting} where each training example is prepended with 5-shot demonstrations. While significant performance disparities between \method and FT are observed across the board, the narrowing gap suggests the growing potential of \method as the underlying LLM continues to evolve. Moreover, fine-tuned small LLMs consistently outperform the few-shot GPT-4, which is the best-performing generalist model on the ProScript dataset. This underscores the continued efficacy of FT in building specialized models, even in the era of LLMs.

\subsection{Further Studies on \method}
\label{sec:ablation_study}
We conduct ablation studies using \textsc{Llama3-8B}, to explore the effect of the few-shot demonstrations and the recounted reference narratives.

\begin{figure}[t]
    \centering
    \includegraphics[width=0.38\textwidth]{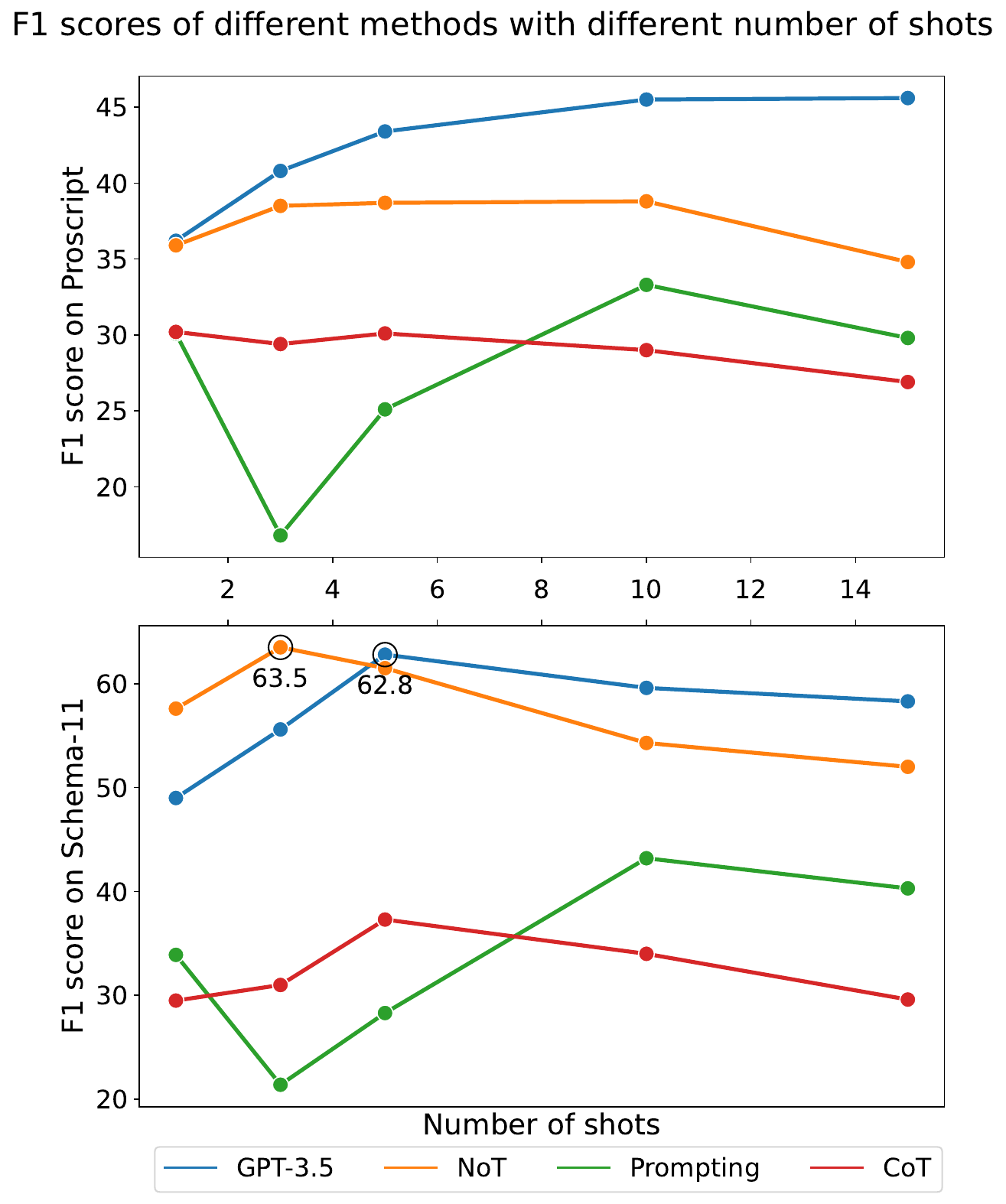}
    \vspace{-2mm}
    \caption{F1 scores on ProScript and Schema-11 in relation to the number of shots in demonstrations. We identify the \textbf{instability} in the standard prompting, and the \textbf{performance plateau} after 5 shots.
    }
    \label{fig:shots}
    \vspace{-4mm}
\end{figure}

\paragraph{Does the number of shots matter?} Figure~\ref{fig:shots} illustrates how F1 scores change with the number of shots in demonstrations. 
As can be seen, GPT-3.5 and \method show resilience to changes in shot numbers after an initial sharp increase. The performance nearly stabilizes in the range of 5-10 shots, though a slight drop is observed later, presumably due to insufficient capability of long-context comprehension \citep{long-context-1, long-context-2}.  Of particular interest is the performance of \method with 3 shots on Schema-11, outperforming the best variant of GPT-3.5 (F1 of 63.5 vs. 62.8). This further illustrates \method's potential of boosting small LLMs in the long run.
It is also noticeable that F1 scores of the standard prompting technique have a V-shape between 1-shot and 5-shot, highlighting its sensitiveness to in-context demonstrations.

We also display the GED scores in relation to number of shots in \Cref{fig:shots_GED}. We observe similar instability in the standard prompting technique, along with the performance plateau after 5 shots.

\paragraph{What characteristics define effective reference narratives?}
Given that reference narratives in \method are machine-generated, we aim to explore what qualities matter most for the TGG task. Here, the three variables influencing reference narratives are: (1) narrative generation model  (GPT-3.5 vs. GPT-4), (2) input format (alphabetical vs. descriptive), and (3) 4 meta prompt types (varying degrees of factuality and readability).
We show detailed meta prompts in \Cref{appx:meta_prompt}. 

\begin{figure}[t]
    \centering
    \includegraphics[width=0.42\textwidth]{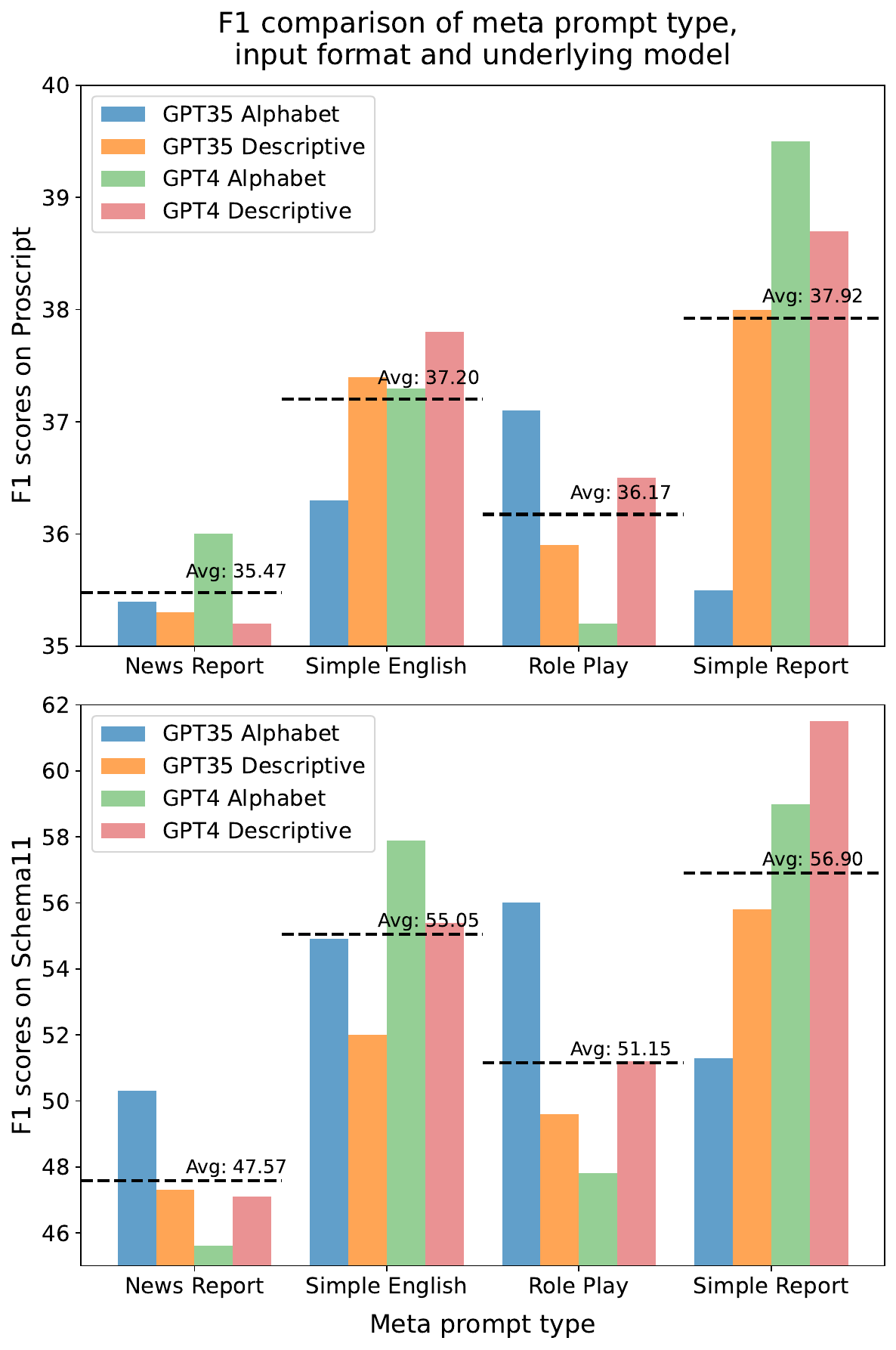}
    \vspace{-2mm}
    \caption{
    F1 scores on ProScript and Schema-11 with different meta prompts. Average performance grouped by prompt type is also shown. Notably, using a \textit{Simple Report}-style, GPT-4 generated narratives lead to the best score due to its \textbf{conciseness}, \textbf{simplicity} and \textbf{factuality}, which are essential qualities for a \textit{high-quality} reference narrative.
    }
    \label{fig:types}
    \vspace{-4mm}
\end{figure}

Figure~\ref{fig:types} and \Cref{fig:type_GED} show results of F1 and GED with varying meta prompts.
Surprisingly, the choice of the generator does not significantly impact the graph quality, with average F1 scores of 36.4 for GPT-3.5 and 37.0 for GPT-4, and GED scores of 1.90 vs. 1.94.
Similarly, there is no significant difference between alphabetical and descriptive input formats.
The most \textit{impactful} factor is the meta prompt type. Grouping performance bars by prompt type reveals a clear variance in model performance. Among the first three groups, \textit{Simple English} narratives, i.e., good for 10-year-olds, stand out. This suggests that narratives should be simple and concise, as verbose ones are less effective. 
We find that \textit{News Report} narratives prioritize procedural and factual content, minimizing distractions like descriptive settings or figurative language that can often be found in both fiction or non-fiction stories.  We thus combine \textit{Simple English} and \textit{News Report} to leverage their strengths, dubbed \textit{Simple Report}. In summary, we identify three key characteristics for quality reference narratives: \textit{conciseness}, \textit{simplicity} and \textit{factuality}.

\paragraph{How faithful is the temporal graph to intermediate narratives?}
Here, we look into whether \method-augmented LLMs are \textbf{self-faithful}, i.e., whether the narrative and the temporal graph \textbf{align} in terms of the temporal order of events. Higher self-faithfulness is crucial and desired, as misalignment would diminish the effort of generating a temporally grounded narrative.\footnote{Faithfulness $\neq$ correctness. A faithful temporal graph may still contain logical errors from the generated narratives.}

Motivated by the recent success of using LLMs as judges \citep{llm-as-judge, Zhang2024ULTRAUL}, we employ GPT-4 to assess the self-faithfulness of 600 randomly sampled outputs by \method-augmented \textsc{Llama3-8B}. We prompt GPT-4 to perform a 5-way assessment and provide judgment rationales. Additionally, GPT-4 is instructed to count the temporal links in the temporal graphs and identify aligned temporal links for a sanity check. This helps humans capture the failure modes and make necessary interventions. Based on automated responses and on-demand human inspections, we find a medium-to-high alignment of 72.8\%. Details of templates and the inspection process are included in \Cref{appx:faithfulness}.

%% file: tables/FT_results.tex
\begin{table}[t]
\centering
\resizebox{0.85\linewidth}{!}{%
\begin{tabular}{lrrrrrr} \toprule
                       & \multicolumn{2}{c}{\textsc{Gemma-7B}}   & \multicolumn{2}{c}{\textsc{Mistral-7B}} & \multicolumn{2}{c}{\textsc{Llama3-8B}}       \\
                       \cmidrule(lr){2-3}
  \cmidrule(lr){4-5}
   \cmidrule(lr){6-7}
                       & F1$\uparrow$ & GED$\downarrow$ & F1$\uparrow$ & GED$\downarrow$ & F1$\uparrow$ & GED$\downarrow$ \\ \midrule
\method & 21.8         & 2.48            & 35.4         & 2.14            & 39.5             & 1.87             \\
FT                     & 68.6             &  1.63               &   71.2           &  1.38               & 71.9                 &  1.40      \\ \bottomrule         
\end{tabular}
}
\vspace{-2mm}
\caption{Performance comparison between \method and fine-tuning (FT) on ProScript. GPT-4 achieves 63.9 F1 and 1.64 GED, both lagging behind fine-tuned LLMs.}
\label{tbl:FT_results}
\vspace{-4mm}
\end{table}

%% file: 6_conclusion.tex
\section{Conclusion}

In this paper, we assess the inherent, global temporal reasoning capabilities of LLMs, by studying the core challenge of temporal reasoning---temporal graph generation (TGG).  
To this end, we propose \methodFull (\method), a novel prompting technique tailored for temporal reasoning. Concretely, with few-show narrative-aware demonstrations as references, \method prompts LLMs to first generate a temporally grounded narrative and then sort the input events topologically into a temporal graph, by manipulating the generation in code space. Extensive experiments showcase \method's effectiveness, demonstrated by its superior performance over GPT-3.5 on multiple metrics, as well as the compatibility of \method with various LLMs.

%% file: Limitations.tex
\section*{Limitations}

\paragraph{Evaluation benchmarks.} In this work, we have included three evaluation benchmarks, aiming to cover a diverse array of genres and domains. Yet, these three benchmarks cannot comprehensively represent the entire spectrum. For example, healthcare and biomedical \citep{bio-temporal} domains offer great opportunities to study temporal graph generation as well.  In future research, we plan to extend \method to more applications, and examine its true generalizability in the wild.  

\paragraph{Human baseline comparison.} The last finding we deliver in \cref{sec:main_results} might not hold for all benchmarks, as the human baseline comparison was conducted solely on the ProScript dataset. We will continue the endeavor of seeking participants to perform human evaluations on the other two datasets to enhance the credibility of our claim.

\paragraph{Scaling effect.} While we recognize the value of investigating models of different sizes to explore the scaling effect of \method, we did not pursue this for two reasons. First, one primary goal is to enable small LLMs (<10B parameters) to match the performance of larger ones like GPT-3.5/4 (RQ2). Second, among the three base models selected in this work, the open-weight Mistral only has a 7B version; while Gemma does have 2B and 7B versions, preliminary results showed that the 2B version yielded subpar performance (e.g., poor instruction-following, outputs are simply concatenations of events in the input order). As for LLAMA3 (8B vs. 70B), we couldn't produce results for 70B due to computational constraints.

\paragraph{GPU resources.} The base LLMs used in this work are of 7 to 8 billions parameters. 
It is thus more time-consuming than traditionally small models like BERT \citep{devlin-etal-2019-bert} at inference time, which in turn results in a higher carbon footprint.
Specifically, we run each base LLM on 1 single NVIDIA A40 or NVIDIA L40 with significant CPU and memory resources. The combined inference time for each LLM on the three benchmarks ranges from 10 to 20 hours, depending on the configurations.

%% file: Appendix.tex
\input{tables/full_results}

\clearpage

\section{Additional Implementation Details}
\label{appx:implementation}
\paragraph{Few-shot Demonstration Selection.}
To construct the demonstration bank, we select 15 examples from the training set of ProScript, following \citet{MadaanTGHGW0DPY23}. We do so because we expect to include non-linear temporal graph examples in our demonstrations, for which only ProScript can fulfill the requirement. Then, we use the same demonstrations as few-shot examples for experiments, regardless of the evaluation benchmark. 

\paragraph{Model Cards.} In this work, we have experimented with 3 base LLMs. Below lists the exact Huggingface model cards used in this work.

\begin{itemize}[leftmargin=2em,itemsep=0em,topsep=1pt,parsep=1pt,partopsep=1pt]
    \item \textsc{Gemma-7B}: \texttt{google/gemma-7b-it}
    
    \item \textsc{Mistral-7B}: \texttt{mistralai/Mistral-7B-Instruct-v0.2}
    
    \item \textsc{Llama3-8B}: \texttt{meta-llama/Meta-Llama-3-8B-Instruct}
\end{itemize}

\section{Dataset Processing}
\label{appx:dataset_processing}
This section documents the processing steps performed on Schema-11 and WikiHow Script to cater for the temporal reasoning task of our interest. We do not use any Python packages for dataset processing. Meanwhile, based on our inspection, we do not spot any offensive content in these three datasets.

\paragraph{Schema-11.} In their original annotations, an event node is marked in arg$_0$-trigger-arg$_1$ format, and we manually convert it to a natural sentence. We specifically adopt annotations under \textit{schemas\_dan\_d} directory.

\paragraph{WikiHow Script corpus.} The original dataset features multilingualism, while we only take their English portion for this study. Then, We only keep ordered how-to articles where steps are presented in chronological order. Lastly, we cap the maximum number of steps at 20, which reduces the corpus size from $3,3035$ to $2,077$.

\section{Complete Examples}
\label{appx:outputs}

\begin{figure}[t]
    \centering
    \includegraphics[width=0.4\textwidth]{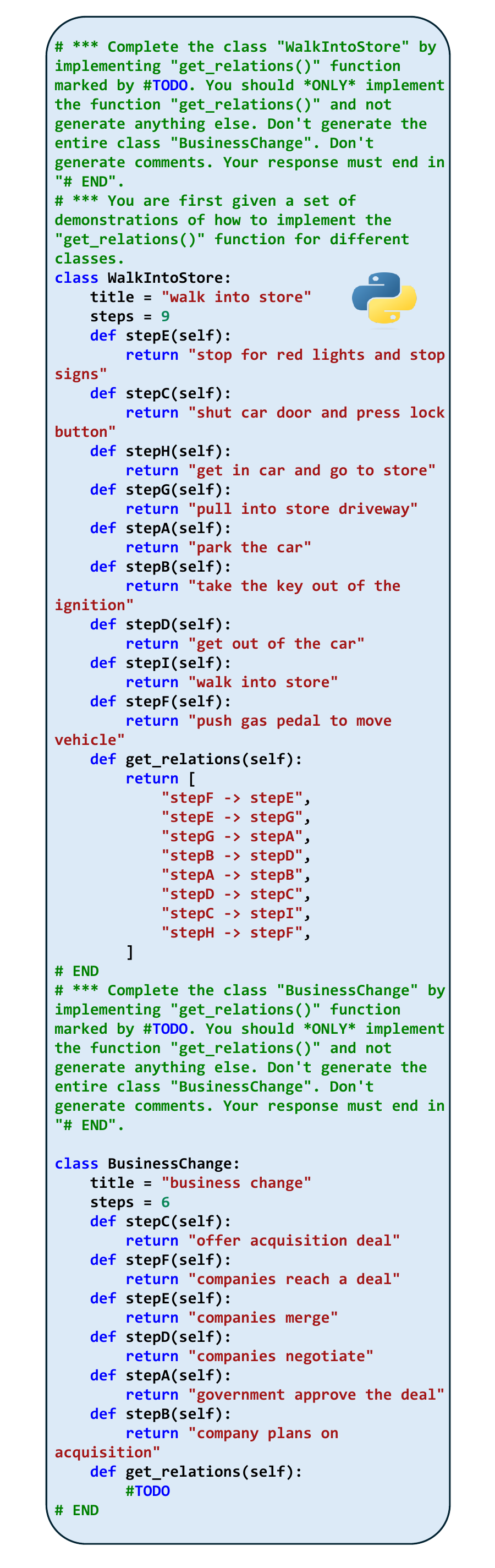}
    \caption{ Input for Standard Prompting with 1-shot demonstration. The input for CoT is almost identical to this one, except for an additional comment ``Let's think step by step'' added right above \texttt{get\_relations(self)}  }
    \vspace{-4mm}
    \label{fig:full_example}
\end{figure}

\begin{figure}[t]
    \centering
    \vspace{-20mm}
    \includegraphics[width=0.38\textwidth]{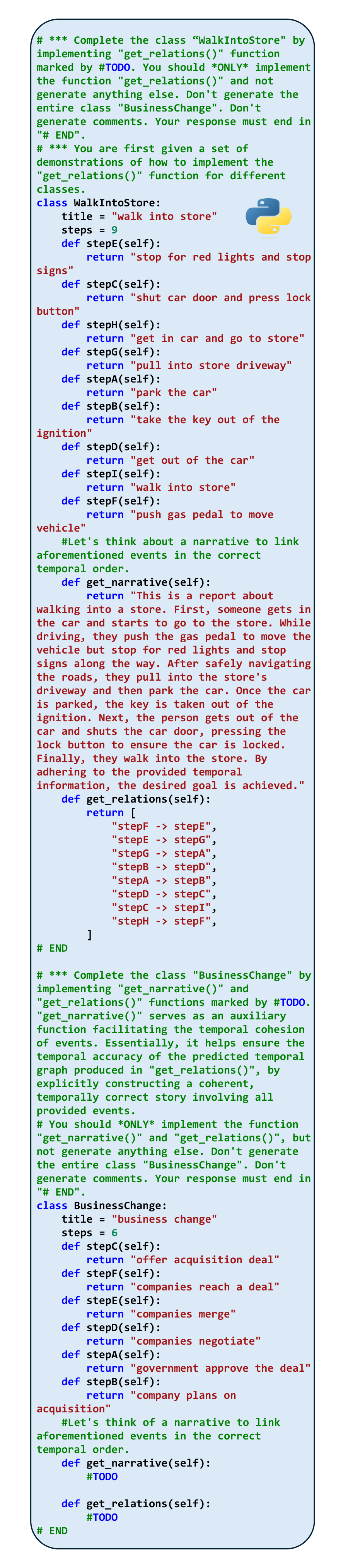}
    \caption{ Input for \method with 1-shot demonstration including a high-quality reference narrative.}
    \vspace{-4mm}
    \label{fig:full_example_gensort}
\end{figure}

\begin{figure}[t]
    \centering
    \includegraphics[width=0.4\textwidth]{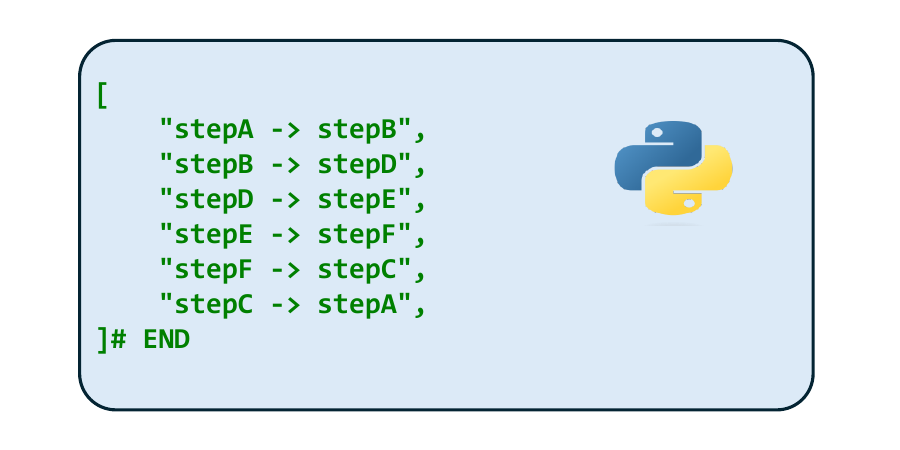}
    \caption{ Output by Standard Prompting.  }
    \vspace{-4mm}
    \label{fig:output_standard_prompting}
\end{figure}

\begin{figure}[t]
    \centering
    \includegraphics[width=0.4\textwidth]{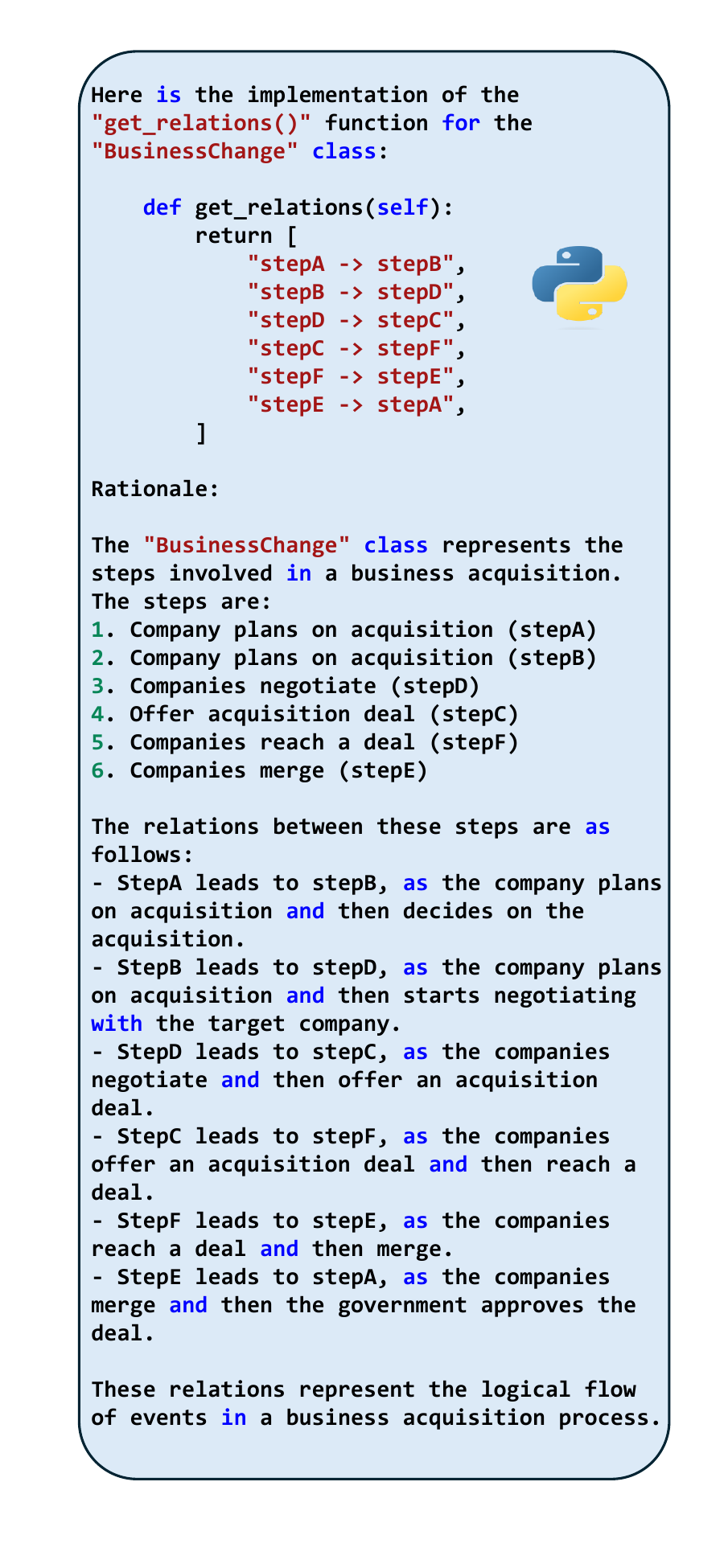}
    \caption{ Output by CoT.  }
    \vspace{-4mm}
    \label{fig:output_CoT}
\end{figure}

\begin{figure}[t]
    \centering
    \includegraphics[width=0.4\textwidth]{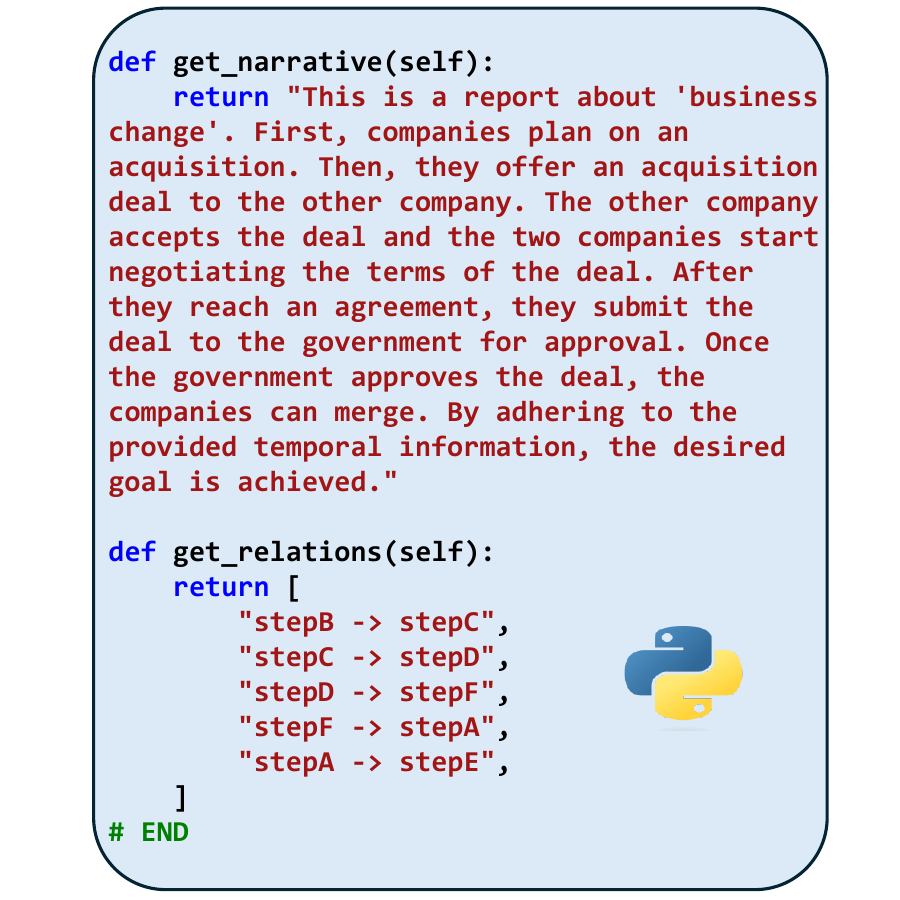}
    \caption{ Output by \method.  }
    \vspace{-4mm}
    \label{fig:output_gensort}
\end{figure}

Using the same example as in \Cref{fig:task} and \Cref{fig:architecture}, we show the complete examples (including generations by one base LLM, \textsc{Llama3-8B}) of Standard Prompting, CoT and \method. We first show the input part of Standard Prompting and CoT in \Cref{fig:full_example}, and the input of \method in \Cref{fig:full_example_gensort}. Outputs by Standard Prompting, CoT and \method are displayed in \Cref{fig:output_standard_prompting},  \Cref{fig:output_CoT} and \Cref{fig:output_gensort}, respectively. As we can easily see, the output of Standard Prompting is completely wrong and fails to capture any correct temporal relation. Worse still, it even forms a loop. For the output of CoT, at least, it gets one temporal relation correct. However, the generated rationales are verbose, not to-the-point, and the mixture of natural language and programming language in the output might confuse the generation process as well. In contrast, the generated temporal graph by \method captures most of the right temporal relations, yielding a high F1 score of 80 points, and a very low GED, which is just 1.

\section{Meta Prompt}
\label{appx:meta_prompt}

\input{tables/meta_prompt_type}

\input{tables/faithfulness_check_template}

\begin{figure}[t]
    \centering
    \includegraphics[width=0.4\textwidth]{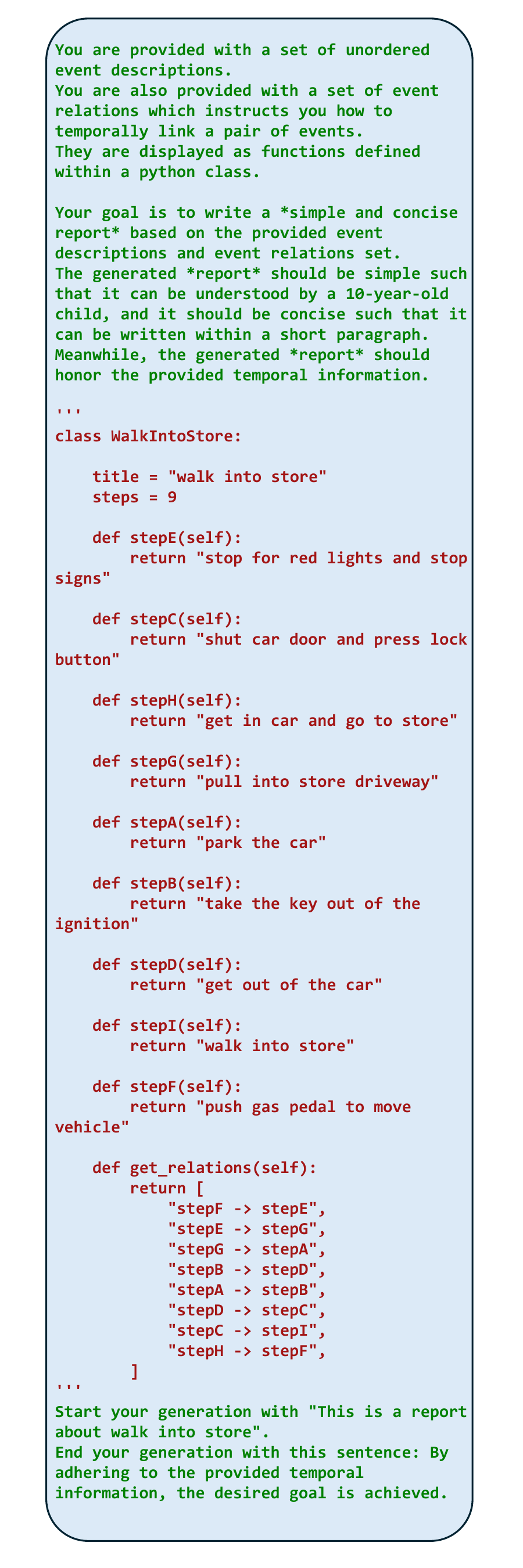}
    \caption{ Meta prompt used to generate reference narrative, where the input format \textbf{alphabetical} and the meta prompt type is \textit{Simple Report}.  }
    \vspace{-4mm}
    \label{fig:meta_alphabetical}
\end{figure}

\begin{figure}[t]
    \centering
    \includegraphics[width=0.4\textwidth]{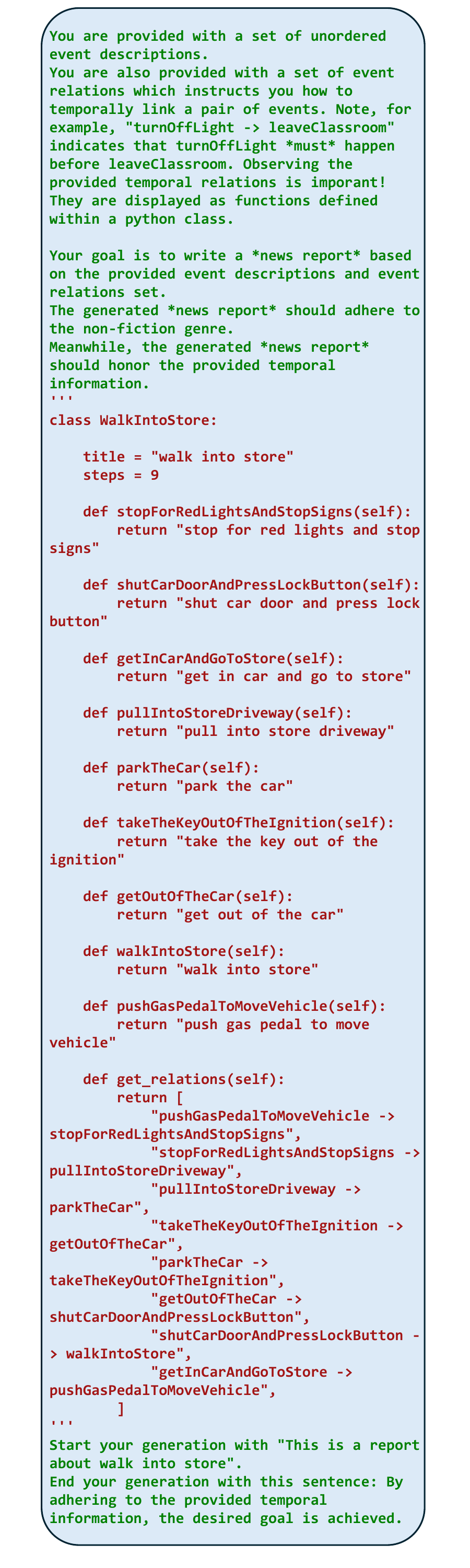}
    \caption{ Meta prompt used to generate reference narrative, where the input format \textbf{descriptive} and the meta prompt type is \textit{News Report}.  }
    \vspace{-4mm}
    \label{fig:meta_descriptive}
\end{figure}

This section discusses the major components of a meta prompt, used to generate reference narratives. As shown in \Cref{fig:meta_alphabetical} and \Cref{fig:meta_descriptive}, a meta prompt consists of two parts: input (in Python programming language) and instruction (above and below the input). The input contains both $\mathcal{V}$ (event set) and $\mathcal{E}$ (temporal relation set), and the goal is to prompt LLMs to generate a high-quality \textit{reference narrative}. The input has two formats: \textbf{alphabetical} (\Cref{fig:meta_alphabetical}) format where the function header is represented in the same fashion as in \Cref{fig:architecture}, and \textbf{descriptive} (\Cref{fig:meta_descriptive}) where the function header is the camel-cased version of the complete event description. The instruction part specifies how LLMs are supposed to carry out the narrative generation, reflecting different types and genres. Specifically, we designed four different instructions, listed in \Cref{tbl:meta_prompt_type}. They are \textit{News Report}, \textit{Simple English}, \textit{Role Play} and \textit{Simple Report}, which is essentially a seamless combination of  \textit{News Report} and \textit{Simple English}.

\begin{figure}[]
    \centering
    \includegraphics[width=0.4\textwidth]{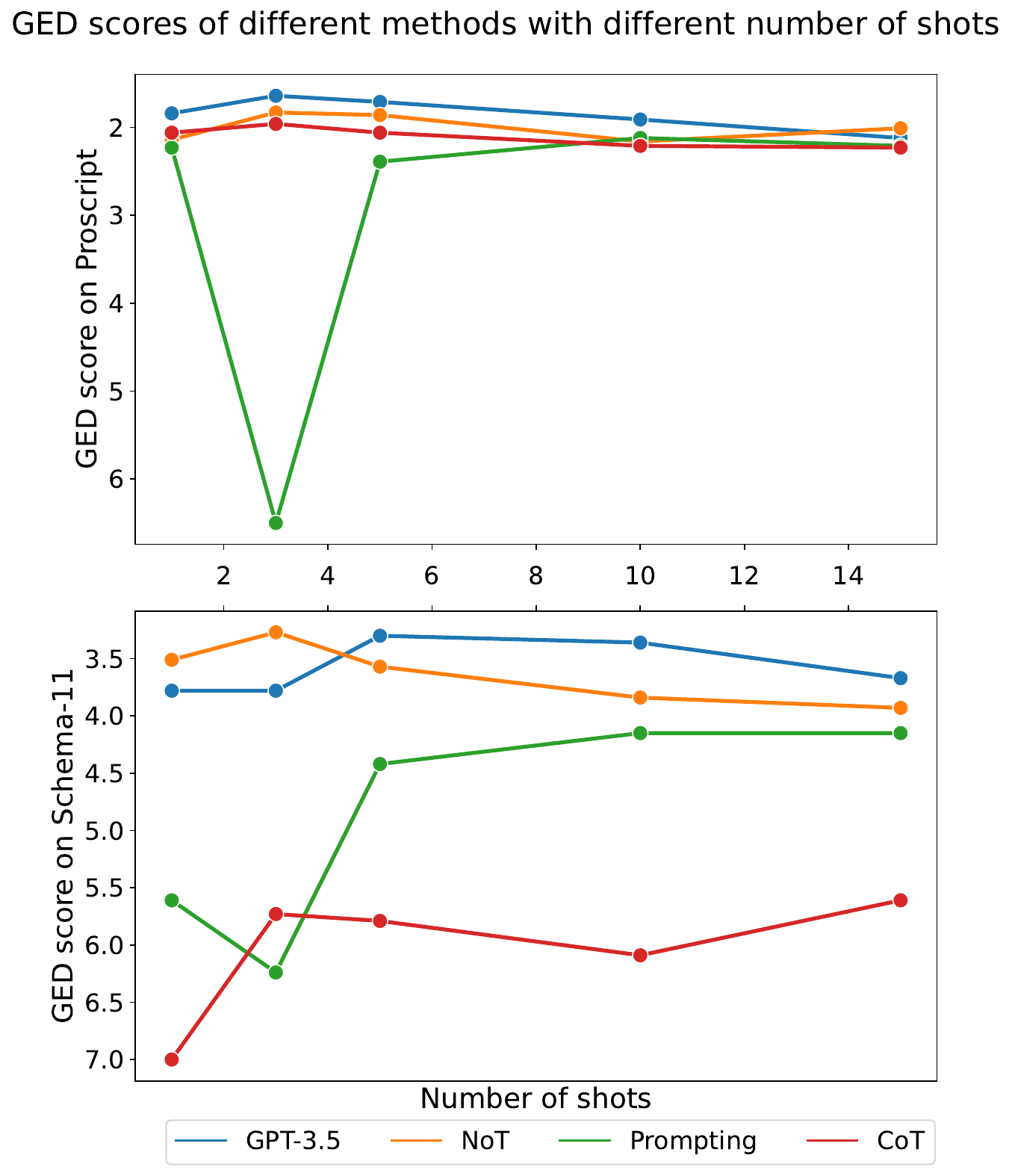}
    \caption{GED scores on ProScript (top) and Schema-11 (bottom) in relation to the number of shots in demonstrations. We identify the \textbf{instability} in the standard prompting, and the \textbf{performance plateau} after 5 shots, along with a slight decline with even more shots.
    }
    \label{fig:shots_GED}
    \vspace{-4mm}
\end{figure}

\begin{figure}[]
    \centering
    \includegraphics[width=0.4\textwidth]{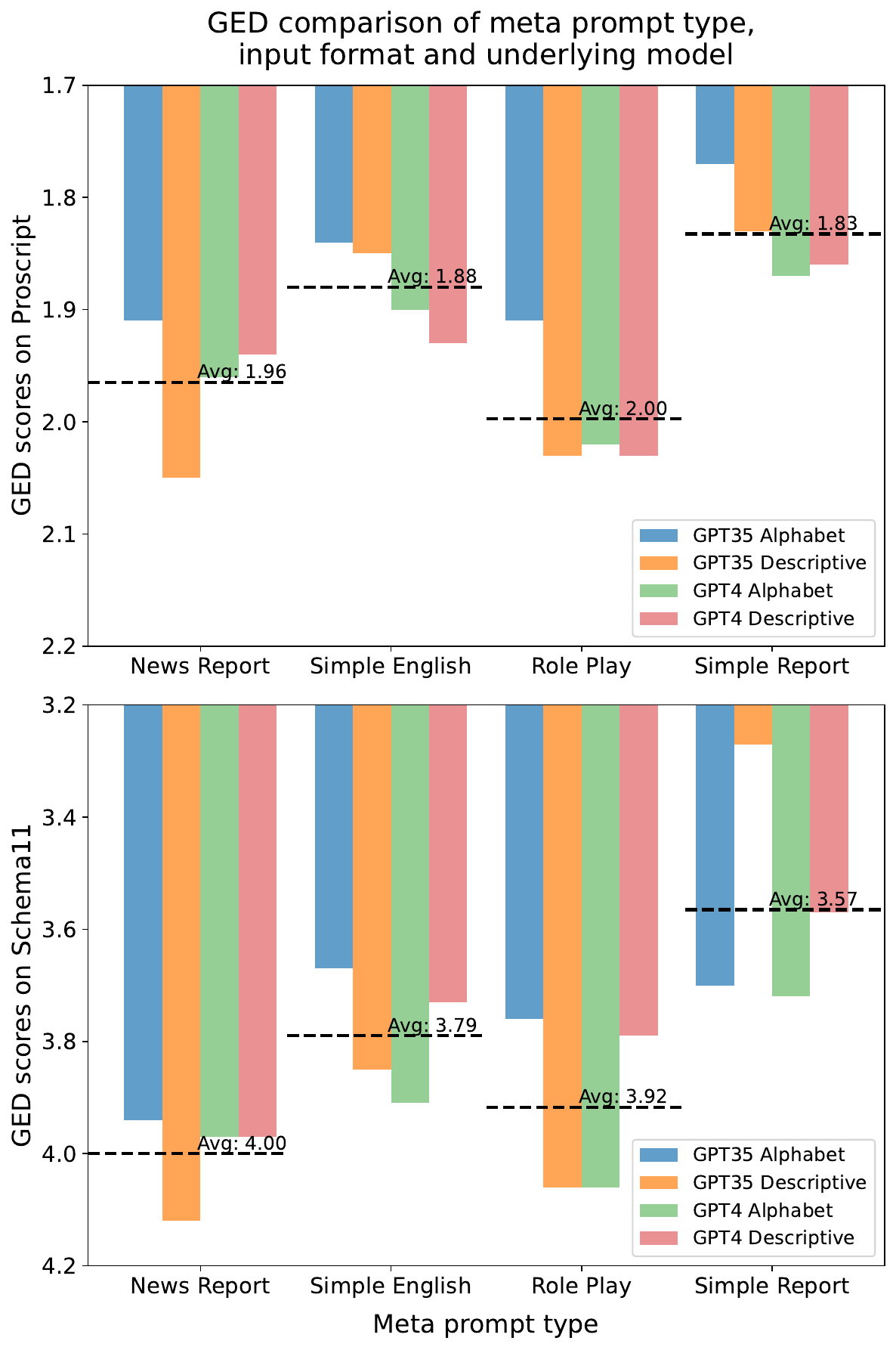}
    \caption{ GED scores on ProScript (top) and Schema-11 (bottom) with different meta prompts. Notably, a \textit{Simple Report}-style, GPT-4 generated narrative leads to the best performance due to its \textbf{conciseness}, \textbf{simplicity} and \textbf{factuality}, which are essential qualities of a \textit{high-quality} reference narrative.
    }
    \label{fig:type_GED}
    \vspace{-4mm}
\end{figure}

\section{Correlation Analysis}
\label{sec:correlation_analysis}
We start our empirical analysis by presenting performances on well-regarded coding benchmarks, HumanEval and MBPP, of the three selected base LLMs included in this work.  Based on these results in \Cref{tbl:code_Eval}, the ranking is Mistral (1) < Gemma (2) < LLAMA3 (3), with the numbers in parentheses indicating their relative scores in this comparison.

\input{tables/code_eval}

Second, regarding instruction-following capability in code completion, we evaluated how well the models adhered to provided instructions (i.e., implementing the return statement of ``get$\_$relations(self)''; Also see \Cref{fig:full_example}). Model outputs are provided in \Cref{tbl:gemma}, \Cref{tbl:llama3} and \Cref{tbl:mistral}.
As can be seen, Mistral produces perfect outputs, while Gemma and LLAMA3 generate the entire class despite being explicitly instructed not to do so. Additionally, Gemma includes a lead phrase, which is also discouraged in the instruction. Therefore, the empirical ranking is Gemma (1) < LLAMA3 (2) < Mistral (3). 

Combining these assessments, the overall ranking for code completion capability is Gemma (3) < Mistral (4) < LLAMA3 (5). This exactly aligns with their performance on our TGG task, suggesting a \textbf{strong positive correlation} between coding capabilities and temporal reasoning. We will continue to explore this relationship through more systematic and quantitative analyses in future work.

\section{Faithfulness Checking Details}
\label{appx:faithfulness}
\Cref{tbl:faithfulness} shows the template being used to prompt GPT-4 to produce a judgment.  GPT-4 performs a 5-way assessment: yes, largely yes, ambivalent, largely no, and no, where yes means exact alignment while no means no alignment at all. With the counting puzzle as a sanity check, we find that GPT-4 does not count the number of temporal links wrong at all. We thus rely on the returned value of \textit{correct temporal links} as a means to determine the failure mode. Before human inspection, the distribution among yes/largely yes/largely no/no is 243/190/32/135, where GPT-4 does not output ``ambivalent''.

\paragraph{Faithfulness Checking Manual Inspection.}
We notice that there are 39 cases where the value of correct temporal links is 0, and 5 cases where GPT-4 refuses to produce a value. Thus, we manually look into these 44 cases. Among these 44 cases, we correct 4 of them. In one case, GPT-4's rationale is ``Additionally, all other links, despite being in the correct order, are rendered incorrect due to the initial incorrect link.'' and GPT-4 marks 0 correct temporal links. However, as GPT-4 has discovered, all except for one link are actually correct, so we change the label from ``no'' to ``yes''. There are three cases where GPT-4 is not judging the faithfulness but instead the \textit{correctness}. As we have noted in the main content, faithfulness is not the same as correctness. For example, one rationale is ``Given the fundamental logical error in the sequence of dialing and answering, all links are considered incorrect in the context of real-world logic, despite matching the narrative's order'' where the narrative mistakenly says ``dialing the phone'' happens after ``answer the phone'', so GPT-4 marks ``no''. Yet, as GPT-4 has also discovered that the temporal graph actually perfectly matches the generated narrative, we thus correct the label from ``no'' to yes. The aforementioned two cases are the ones where GPT-4 got stuck in this assessment task. 

After human inspection, the final adjudicated distribution is 247/190/32/131. This leads to an alignment level of 72.8\% where we consider both ``yes'' and ``largely yes'' as entailing \textit{alignment}.

\clearpage

\begin{table*}[htbp]
    \centering
    \begin{adjustbox}{max width=0.9\textwidth}
    \begin{tabular}{|c|}
        \hline
        \begin{lstlisting}[language=Python, basicstyle=\ttfamily, frame=none]
Sure, here is the completed "get_relations()" function for the "BombingAttacks" class:

class BombingAttacks:

    title = "bombing attacks"
    steps = 10

    def stepA(self):
        return "people are killed"

    def stepF(self):
        return "person places bomb"

    def stepE(self):
        return "bomb explodes"

    def stepG(self):
        return "calls paramedics"

    def stepI(self):
        return "people are taken to hospital"

    def stepJ(self):
        return "paramedics arrive"

    def stepC(self):
        return "paramedics treat injured people"

    def stepH(self):
        return "person detonates bomb"

    def stepD(self):
        return "people are injured"

    def stepB(self):
        return "person plans an attack"

    def get_relations(self):
        return [
            "stepB -> stepE",
            "stepE -> stepJ",
            "stepJ -> stepC",
            "stepC -> stepG",
            "stepG -> stepI",
            "stepI -> stepH",
            "stepH -> stepA",
            "stepA -> stepD",
        ]
        \end{lstlisting}
        \\ \hline
    \end{tabular}
    \end{adjustbox}
    \caption{\textsc{Gemma-7B}'s output.}
    \label{tbl:gemma}
\end{table*}

\clearpage

\begin{table}[htbp]
    \hspace{-13mm}
    \begin{adjustbox}{max width=0.65\textwidth}
    \begin{tabular}{|c|}
        \hline
        \begin{lstlisting}[language=Python, basicstyle=\ttfamily, frame=none]
class BombingAttacks:

    title = "bombing attacks"
    steps = 10

    def stepA(self):
        return "people are killed"

    def stepF(self):
        return "person places bomb"

    def stepE(self):
        return "bomb explodes"

    def stepG(self):
        return "calls paramedics"

    def stepI(self):
        return "people are taken to hospital"

    def stepJ(self):
        return "paramedics arrive"

    def stepC(self):
        return "paramedics treat injured people"

    def stepH(self):
        return "person detonates bomb"

    def stepD(self):
        return "people are injured"

    def stepB(self):
        return "person plans an attack"

    def get_relations(self):
        return [
            "stepB -> stepH",
            "stepH -> stepE",
            "stepE -> stepD",
            "stepD -> stepC",
            "stepC -> stepJ",
            "stepJ -> stepG",
            "stepG -> stepI",
            "stepI -> stepA",
            "stepF -> stepH",
        ]
        \end{lstlisting}
        \\ \hline
    \end{tabular}
    \end{adjustbox}
    \caption{\textsc{Llama3-8B}'s output.}
    \label{tbl:llama3}
\end{table}

\begin{table}[htbp]
    \hspace{13mm}
    \begin{adjustbox}{max width=0.75\textwidth}
    \begin{tabular}{|c|}
        \hline
        \begin{lstlisting}[language=Python, basicstyle=\ttfamily, frame=none]
return [
    "stepB -> stepF",
    "stepF -> stepH",
    "stepH -> stepA",
    "stepH -> stepD",
    "stepD -> stepG",
    "stepG -> stepJ",
    "stepJ -> stepC",
    "stepC -> stepI",
    "stepI -> stepA",
]
        \end{lstlisting}
        \\ \hline
    \end{tabular}
    \end{adjustbox}
    \caption{\textsc{Mistral-7B}'s output.}
    \label{tbl:mistral}
\end{table}

%% file: tables/full_results.tex
\begin{sidewaystable*}
\resizebox{1.05\linewidth}{!}{%
\hspace{-10mm}
\begin{tabular}{lrrrrrrrrrrrrrrrrrrrrrrrrr} \toprule
\multicolumn{1}{c}{\multirow{2}{*}{Method}} & \multicolumn{7}{c}{Proscript}                                                                                 & \multicolumn{7}{c}{Schema-11}                                                                                       & \multicolumn{7}{c}{WikiHow Script}                                                                            & \multicolumn{4}{c}{Avg}                                        \\
 \cmidrule(lr){2-8}
  \cmidrule(lr){9-15}
   \cmidrule(lr){16-22}
    \cmidrule(lr){23-26} 
\multicolumn{1}{c}{}                        & P             & R             & F1            & GED           & $\frac{|\hat{\mathcal{E}}|}{|\mathcal{E}|}$         & $k(\mathcal{G})$          & Cons.         & P             & R             & F1            & GED                 & $\frac{|\hat{\mathcal{E}}|}{|\mathcal{E}|}$         & k(G)          & Cons.         & P             & R             & F1            & GED           & $\frac{|\hat{\mathcal{E}}|}{|\mathcal{E}|}$         & $k(\mathcal{G})$         & Cons.         & F1            & GED           & $k(\mathcal{G})$          & Cons.         \\ \midrule
\multicolumn{26}{c}{Baselines}                                                                                                                                                                                                                                                                                                                                                                                                                                    \\ \midrule
Random                                      & 14.6          & 13.6          & 14.0          & 1.47          & 0.93          & 1.00          & 7.8           & 20.2          & 18.8          & 19.4          & 3.91                & 0.96          & 1.00          & 7.8           & 14.2          & 14.2          & 14.2          & 0.06          & 1.00          & 1.00          & 8.8           & 15.9          & 1.81          & 1.00          & 8.1           \\
GPT-3.5 (0-shot)                            & 18.8          & 18.1          & 18.4          & 2.25          & 0.95          & 1.06          & 38.6          & 30.8          & 30.1          & 30.1          & 4.48                & 1.02          & 1.27          & 30.2          & 17.0          & 17.8          & 17.2          & 2.80          & 1.04          & 1.11          & 40.8          & 21.9          & 3.18          & 1.15          & 36.5          \\
GPT-3.5                                     & 44.9          & 42.3          & 43.4          & 1.71          & 0.92          & 1.07          & 38.8          & 65.8          & 60.5          & 62.8          & 3.30                & 0.92          & 1.36          & 50.2          & 31.0          & 31.1          & 31.0          & 1.58          & 1.01          & 1.10          & 35.4          & 45.7          & 2.20          & 1.18          & 41.5          \\
GPT-4                                       & 65.7          & 62.6          & 63.9          & 1.64          & 0.94          & 1.02          & 61.4          & 44.9          & 43.5          & 44.1          & 7.97                & 0.57          & 0.64          & 46.3          & 43.0          & 43.1          & 43.0          & 1.71          & 0.98          & 1.04          & 48.5          & 50.3          & 3.77          & 0.90          & 52.1          \\ \midrule
\multicolumn{26}{c}{\textsc{Gemma-7B} \citep{gemma}}                                                                                                                                                                                                                                                                                                                                                                                                                                     \\ \midrule
Standard Prompting                          & 20.2          & 19.4          & 19.7          & \textbf{2.35} & 0.96          & 1.02          & \textbf{20.4} & 28.5          & 27.6          & 27.8          & 5.03                & 0.97          & \textbf{1.03} & 18.3          & 16.9          & 18.5          & 17.5          & \textbf{2.88} & \textbf{0.99} & \textbf{0.96} & \textbf{17.3} & 21.7          & \textbf{3.42} & \textbf{1.00} & \textbf{18.7} \\
Chain-of-Thought                            & 20.3          & 19.9          & 20.0          & \textbf{2.35} & \textbf{0.98} & 1.01          & 20.0          & 26.4          & 26.4          & 26.4          & 5.03                & \textbf{1.01} & \textbf{1.03} & 14.9          & 13.0          & 14.5          & 13.6          & 5.91          & 0.77          & 0.73          & 11.5          & 20.0          & 4.43          & 0.92          & 15.5          \\
\method (no reference)            & 20.5          & 19.7          & 20.0          & 2.47          & 0.95          & \textbf{1.00} & 17.3          & 28.6          & 27.4          & 27.9          & \textbf{4.78}       & 0.96          & 1.09          & 18.1          & 14.6          & 16.1          & 15.2          & 5.03          & 0.80          & 0.81          & 13.9          & 21.0          & 4.09          & 0.97          & 16.4          \\
\method (alphabetical meta)              & \textbf{22.4} & \textbf{21.4} & \textbf{21.8} & 2.48          & 0.95          & \textbf{1.00} & 18.3          & \textbf{36.9} & \textbf{35.5} & \textbf{36.0} & 4.84                & 0.95          & 1.06          & 19.7          & 17.1          & \textbf{18.9} & \textbf{17.9} & 2.95          & 0.96          & \textbf{0.96} & 16.9          & \textbf{25.2} & \textbf{3.42} & 1.01          & 18.3          \\
\method (descriptive meta)          & 21.9          & 21.0          & 21.3          & 2.60          & 0.94          & 0.99          & 17.8          & 35.0          & 34.8          & 34.8          & 5.00                & 0.98          & 1.06          & \textbf{20.8} & \textbf{17.2} & \textbf{18.9} & \textbf{17.9} & \textbf{2.88} & 0.96          & 0.95          & 16.8          & 24.7          & 3.49          & \textbf{1.00} & 18.5          \\ \midrule
\multicolumn{26}{c}{\textsc{Mistral-7B} \citep{mistral}}                                                                                                                                                                                                                                                                                                                                                                                                                                   \\ \midrule
Standard Prompting                          & 31.7          & 30.0          & 30.7          & 2.16          & 0.94          & 1.05          & 22.3          & 37.6          & 33.7          & 35.3          & 4.55                & 0.93          & 1.12          & 29.1          & \textbf{22.3} & \textbf{23.0} & \textbf{22.5} & \textbf{2.09} & 1.02          & 1.11          & \textbf{18.9} & 29.5          & 2.93          & 1.09          & 23.4          \\
Chain-of-Thought                            & 30.7          & 29.3          & 29.8          & 2.66          & 0.92          & \textbf{1.02} & 22.1          & 36.0          & 34.6          & 35.2          & 5.33                & 0.95          & 0.94          & 30.5          & 20.9          & 20.5          & 20.5          & 2.59          & \textbf{0.99} & 1.10          & 17.4          & 28.5          & 3.53          & \textbf{1.02} & 23.3          \\
\method (no reference)            & 33.4          & 31.9          & 32.5          & 3.04          & 0.89          & 0.95          & 19.4          & 44.0          & 41.3          & 42.3          & 5.27                & 0.86          & \textbf{1.00} & 27.6          & 21.7          & 22.3          & 21.8          & 3.33          & 0.91          & \textbf{0.98} & 15.4          & 32.2          & 3.88          & 0.98          & 20.8          \\
\method (alphabetical meta)              & 35.9          & 34.8          & 35.2          & \textbf{2.11} & \textbf{0.96} & \textbf{1.02} & 22.4          & 52.8          & 49.5          & 50.9          & 4.30                & 0.95          & 1.03          & \textbf{36.1} & 21.4          & 22.2          & 21.7          & 2.49          & 0.96          & 1.04          & 14.8          & 35.9          & 2.97          & 1.03          & \textbf{24.4} \\
\method (descriptive meta)          & \textbf{36.2} & \textbf{34.9} & \textbf{35.4} & 2.14          & \textbf{0.96} & \textbf{1.02} & \textbf{23.0} & \textbf{54.5} & \textbf{51.4} & \textbf{52.7} & \textbf{3.90}       & \textbf{0.96} & 1.06          & 32.5          & 21.8          & 22.5          & 22.1          & 2.53          & 0.95          & 1.04          & 15.1          & \textbf{36.7} & \textbf{2.86} & 1.04          & 23.5          \\  \midrule
\multicolumn{26}{c}{\textsc{Llama-8B} \citep{llama3modelcard}}                                                                                                                                                                                                                                                                                                                                                                                                                                    \\  \midrule
Standard Prompting                          & 27.3          & 23.4          & 25.1          & 2.39          & 0.85          & 1.18          & 19.9          & 30.8          & 26.6          & 28.3          & 4.42                & 0.91          & 1.24          & 19.9          & 21.5          & 20.1          & 20.6          & 1.17          & 0.97          & 1.07          & 21.2          & 24.7          & 2.66          & 1.16          & 20.3          \\
Chain-of-Thought                            & 30.1          & 30.4          & 30.1          & 2.06          & \textbf{1.00} & \textbf{1.00} & 23.3          & 38.0          & 36.9          & 37.3          & 5.79                & 0.83          & 0.85          & 23.5          & 21.9          & 23.5          & 22.6          & \textbf{0.99} & 1.05          & \textbf{1.02}          & \textbf{24.3} & 30.0          & 2.95          & 0.96          & 23.7          \\
\method (no reference)            & 36.7          & 34.7          & 35.5          & 1.88          & 0.93          & \textbf{1.00} & 25.3          & 54.0          & 51.6          & 52.6          &  \textbf{3.18} & \textbf{0.96} & 1.12          & 35.0          & 25.4          & 25.6          & 25.4          & \textbf{0.99} & 0.99          & \textbf{1.02}          & 20.9          & 37.8          & \textbf{2.02} & 1.05          & 27.1          \\
\method (alphabetical meta)              & \textbf{40.4} & \textbf{38.8} & \textbf{39.5} & 1.87          & 0.95          & 1.01          & \textbf{28.8} & 61.9          & 56.8          & 59.0          & 3.72                & 0.93          & 1.12          & 39.1          & 25.9         & \textbf{26.9} & 26.3          & 1.01          & 1.01          & 1.03          & 22.5          & 41.6          & 2.20          & 1.05          & 30.1          \\
\method (descriptive meta)          & 39.8          & 38.0          & 38.7          & \textbf{1.86} & 0.94          & 1.01          & 28.4          & \textbf{64.1} & \textbf{59.6} & \textbf{61.5} & 3.57                & \textbf{0.96} & \textbf{1.09} & \textbf{45.6} & \textbf{26.2} & \textbf{26.9} & \textbf{26.5} & 1.04          & \textbf{1.00} & 1.03          & 22.3          & \textbf{42.2} & 2.16          & \textbf{1.04} & \textbf{32.1} \\ \bottomrule
\end{tabular}
}
\caption{Full results of three base LLMs and select strong baselines on our compiled suite of TGG evaluation benchmarks. For each base model, best results are in \textbf{bold}. For precision (P), recall (R), F1 and Consistency (Cons.). a higher number indicates a better performance. For GED, a lower number indicates a better results. For $\frac{|\hat{\mathcal{E}}|}{|\mathcal{E}|}$, the optimal value is 1. For $k(\mathcal{G})$, the best value is 1 and numbers smaller than 1 are not favored which indicates LLMs fail to generate a valid graph for some input scenarios.  *Unless otherwise noted, 5-shot demonstrations are provided.}
\label{tbl:full_results}
\end{sidewaystable*}

%% file: tables/meta_prompt_type.tex
\begin{table*}[t]
\resizebox{1.00\linewidth}{!}{%
\begin{tabular}{lllll} \toprule
Instruction Type        & Detailed Instruction          \\ \midrule
\textbf{News Report}    & \begin{tabular}[c]{@{}l@{}}You are provided with a set of unordered event descriptions. \\ You are also provided with a set of event relations which instructs you how to temporally link a pair of events. \\ They are displayed as functions defined within a python class. \textbackslash{}n\\ Your goal is to write a *news report* based on the provided event descriptions and event relations set. \\ The generated *news report* should adhere to the non-fiction genre. \\ Meanwhile, the generated *news report* should honor the provided temporal information. \textbackslash{}n\end{tabular}   \\ \midrule
\textbf{Simple English} & \begin{tabular}[c]{@{}l@{}}You are provided with a set of unordered event descriptions. \\ You are also provided with a set of event relations which instructs you how to temporally link a pair of events. \\ They are displayed as functions defined within a python class. \textbackslash{}n\\ Your goal is to write a *simple and concise story* based on the provided event descriptions and event relations set. \\ The generated *story* should be simple such that it can be understood by a 10-year-old child, \\and it should be concise such that it can be written within a short paragraph.\\ Meanwhile, the generated *story* should honor the provided temporal information. \textbackslash{}n\end{tabular}  \\ \midrule
\textbf{Role Play}      & \begin{tabular}[c]{@{}l@{}}You are provided with a set of unordered event descriptions.\\ You are also provided with a set of event relations which instructs you how to temporally link a pair of events.\\ They are displayed as functions defined within a python class. \textbackslash{}n\\ Your goal is to write a *simple and concise story* based on the provided event descriptions and event relations set.\\ The generated *story* should honor the provided temporal information. \textbackslash{}n\\ Now, imagine you are a character in the *story*. \\ Let's write a *story* that clearly depicts how you, as a character, experience the events, and how you react to them.\end{tabular}   \\ \midrule
\textbf{Simple Report}  & \begin{tabular}[c]{@{}l@{}}You are provided with a set of unordered event descriptions. \\ You are also provided with a set of event relations which instructs you how to temporally link a pair of events. \\ They are displayed as functions defined within a python class. \textbackslash{}n\\ Your goal is to write a *simple and concise report* based on the provided event descriptions and event relations set. \\ The generated *report* should be simple such that it can be understood by a 10-year-old child, \\and it should be concise such that it can be written within a short paragraph.\\ Meanwhile, the generated *report* should honor the provided temporal information. \textbackslash{}n\end{tabular} \\ \bottomrule
\end{tabular}
}
\caption{Detailed instruction for different meta prompt type, a.k.a., instruction type.}
\label{tbl:meta_prompt_type}
\end{table*}

%% file: tables/faithfulness_check_template.tex
\begin{table*}[]
\resizebox{1.0\linewidth}{!}{%
\begin{tabular}{l} \\ \toprule
The temporal graph is represented as a list of tuples, where each tuple contains two events. The first event happens before the second event, connected with '-\textgreater{}'.\textbackslash n \\
Your task is to determine whether the narrative is faithful to the temporal graph.  \\The faithfulness is solely determined by whether the temporal relations in the temporal graph *honor* the chronological order among events in the narrative.\textbackslash n\\
How to make an assessment: If the temporal graph is completely faithful to the narrative, type 'yes'. If largely faithful with minor mistakes, type 'largely yes'. \\If largely not faithful with only a few temporal relations captured, type 'largely no'. If completely not faithful, type 'no'. For other cases, type 'ambivalent'.\textbackslash n\\
Your response should be in the following format:\textbackslash n\textbackslash n \\\\

'''\\
Answer: yes/largely yes/ambivalent/largely no/no\\
Rationale: <your rationale\textgreater{}\\
Temporal links: <count the number of temporal links in the graph\textgreater{}\\
Correct temporal links: <determine the number of *correct* temporal links\textgreater{}\\
'''\\
\\
Let's start!\\
\\
Scenario: [SCENARIO]\\
Events: [EVENTS]\\
Narrative: [NARRATIVE]\\
Temporal Graph: [TEMPORAL GRAPH]      \\ \bottomrule                                                                                                        
\end{tabular}
}
\vspace{-2mm}
\caption{Template used to prompt GPT-4 for self-faithfulness checking. [$\cdot$] are placeholders that will be replaced with real contents to be examined when prompted. <$\cdot$> are also placeholders but are used to instruct GPT-4 what the output format should look like.}
\vspace{-4mm}
\label{tbl:faithfulness}
\end{table*}

%% file: tables/code_eval.tex
\begin{table}[t]
\centering
\resizebox{1.01\linewidth}{!}{%
\begin{tabular}{llll} \toprule
Benchmark          & Gemma-7B & Mistral-7B & Llama3-8B \\  \midrule
HumanEval (0-shot) & 34.1     & 28.0       & 60.4      \\
MBPP (3-shot)      & 51.5     & 50.8       & 67.7      \\
Avg                & 42.8     & 39.4       & 64.1     \\ \bottomrule
\end{tabular}
}
\vspace{-2mm}
\caption{Select LLMs' performance on code generation tasks. Results are taken from \citep{phi3}}
\vspace{-4mm}
\label{tbl:code_Eval}
\end{table}